\documentclass{article}

\usepackage[final, nonatbib]{neurips_2025}

\usepackage[utf8]{inputenc} % allow utf-8 input
\usepackage[T1]{fontenc}    % use 8-bit T1 fonts
\usepackage{hyperref}       % hyperlinks
\usepackage{url}            % simple URL typesetting
\usepackage{booktabs}       % professional-quality tables
\usepackage{amsfonts}       % blackboard math symbols
\usepackage{textcomp}      % support for Unicode minus sign
\usepackage{nicefrac}       % compact symbols for 1/2, etc.
\usepackage{microtype}      % microtypography
\usepackage{amsmath}
\usepackage{graphicx}   % already needed for \includegraphics
\usepackage{subcaption} % defines the subfigure environmen
\usepackage[numbers]{natbib}
\usepackage{todonotes}
\usepackage[percent]{overpic} 
\usepackage{multirow}

\title{Assessing the Geographic Generalization \\ and Physical Consistency of Generative Models \\ for Climate Downscaling}

\author{%
  Carlo Saccardi$^{1,2}$\thanks{Corresponding author: c.saccardi@tudelft.nl} \quad
  Maximilian Pierzyna$^{1}$ \quad
  Haitz Sáez de Ocáriz Borde$^{2,3}$ \quad
  Simone Monaco$^{4}$ \\
  \textbf{Cristian Meo$^{1}$} \quad
  \textbf{Pietro Liò$^{2}$} \quad
  \textbf{Rudolf Saathof$^{1}$} \quad
  \textbf{Geethu Joseph$^{1}$} \quad
  \textbf{Justin Dauwels$^{1}$} \\[1.5ex]
  $^{1}$Delft University of Technology, Delft, The Netherlands \\
  $^{2}$University of Cambridge, Cambridge, United Kingdom \\
  $^{3}$University of Oxford, Oxford, United Kingdom \\
  $^{4}$Politecnico di Torino, Turin, Italy
}

\begin{document}

\maketitle
\begin{abstract}
Kilometer-scale weather data is crucial for real-world applications but remains computationally intensive to produce using traditional weather simulations. An emerging solution is to use deep learning models, which offer a faster alternative for climate downscaling. However, their reliability is still in question, as they are often evaluated using standard machine learning metrics rather than insights from atmospheric and weather physics. This paper benchmarks recent state-of-the-art deep learning models and introduces physics-inspired diagnostics to evaluate their performance and reliability, with a particular focus on geographic generalization and physical consistency. Our experiments show that, despite the seemingly strong performance of models such as CorrDiff, when trained on a limited set of European geographies (e.g., central Europe), they struggle to generalize to other regions such as Iberia, Morocco in the south, or Scandinavia in the north. They also fail to accurately capture second-order variables such as divergence and vorticity derived from predicted velocity fields. These deficiencies appear even in in-distribution geographies, indicating challenges in producing physically consistent predictions. We propose a simple initial solution: introducing a power spectral density loss function that empirically improves geographic generalization by encouraging the reconstruction of small-scale physical structures. The code for reproducing the experimental results can be found at \url{https://github.com/CarloSaccardi/PSD-Downscaling}
\end{abstract}

%------------------------------------------------------------------------------------------------------------------------------
%%%%% Introduction %%%%%%%%%%%
%------------------------------------------------------------------------------------------------------------------------------
\section{Introduction}
Weather forecasts at kilometer-scale resolution are essential for many applications, including energy system planning \citep{QIU2024100263}, agriculture \citep{NeedforHighRes}, and natural disaster preparedness \citep{doi:10.1126/science.1112121}. However, global reanalysis datasets like ERA5 \citep{era5}, commonly used to train models of weather dynamics, are available only at coarse spatial resolutions (e.g., 25 km for ERA5 \citep{era5}). Consequently, global machine learning (ML)-based weather forecasting systems trained on these datasets \citep{Bodnar2025, Sanchez2023, Price2025, Oskarsson2024, Bi2023} are inherently limited to the same coarse scale. This prevents them from capturing fine-scale dynamics and physical processes vital for accurate local impact assessments, leading to systematic discrepancies from real-world measurements \citep{https://doi.org/10.1029/2009RG000314}. 

Traditionally, dynamical downscaling is widely used to address the aforementioned limitation by deriving high-resolution information from coarse global forecasts \citep{salathe2010regional}. This is achieved by numerically integrating the governing equations on a nested, limited-area grid, allowing simulations to capture small-scale physical processes. Still, computational cost grows steeply with domain size and simulation length, making this approach infeasible for large scale and detailed runs~\citep{Bader2009ClimateModels}. Recently, machine learning has been proposed as a faster alternative: statistical downscaling (often called super-resolution in computer vision \citep{harder2023hard}) learns a data-driven coarse-to-fine mapping. Once trained, the models can produce kilometer-scale fields several orders of magnitude faster than traditional dynamical downscaling \cite{Wilby1998}. However, most existing deep learning approaches neither verify whether their outputs are physically consistent nor test whether the learned mapping generalizes to climates and geographic terrains unseen during training. This represents a literature gap as high performance on in-distribution results may not fully reflect their reliability in other conditions.

In this work, we critically evaluate state-of-the-art deep-learning downscalers on physical consistency and generalization and show that physics-informed loss can improve both. Our contributions can be summarized as follows:
\begin{enumerate}
    \item We assess generalization performance by benchmarking three state-of-the-art deep-learning downscalers trained on a Central-European subset and tested on two held-out regions, the Iberia, Morocco region and Northern Scandinavia. We assess geographic transfer using standard machine-learning metrics and observe that the considered models significantly underperform in out-of-distribution (OOD) geographic domains.
    \item Guided by diagnostics from atmospheric-physics literature, we assess physical consistency by examining power spectra of the following second order derived variables: horizontal kinetic energy, mass-continuity/divergence, and relative vorticity. This broadens evaluation beyond conventional ML scores and provides novel insights from atmospheric physics, revealing a lack of physical consistency in the predictions.
    \item We introduce a Power spectral density (PSD) loss that explicitly rewards the reconstruction of small-scale physical dynamics, acts as a frequency-domain regularizer, and alleviates both OOD generalization gaps and physics-consistency shortcomings.
\end{enumerate}

%------------------------------------------------------------------------------------------------------------------------------
%%%%% Related work %%%%%%%%%%%
%------------------------------------------------------------------------------------------------------------------------------
\section{Related work} 
\paragraph{Generalization} Statistical downscaling focuses on learning the conditional distribution $p(\mathbf{q}|\mathbf{x})$ from a set of paired data \(\{(\mathbf{x}_i,\mathbf{q}_i)\}_{i=1}^{N}\), where \(\mathbf{x}\in\mathbb{R}^{C\times h\times w}\) and \(\mathbf{q}\in\mathbb{R}^{c\times H\times W}\) represent the input and target, respectively. Here, \(N\) denotes the number of paired samples, \(C\) is the number of meteorological variables given as input, and \(c\) is the number of meteorological variables to be downscaled. Also, $(h,w)$ and $(H,W)$ denote the height and width of the coarse and fine grids, respectively. Global physics-based simulations are computationally expensive, so any statistical downscaling scheme is useful only if it can be applied beyond the few regions where high-quality training data exist \citep{Glawion2025}. This scarcity demands models that generalize across disparate climates and topographies; otherwise, distribution shifts between the training and target domains can sharply degrade skill and even produce physically implausible outputs. Nevertheless, most prior studies still train and evaluate on the same geography, often limited to a single country \citep{mardani2025}, leaving the cross-climate robustness of ML downscalers largely unexplored.

\paragraph{Physical consistency} Assessing physical consistency is crucial as it is common to diagnose (i.e., derive) one set of variables from the set of prognosed (i.e., directly estimated) variables based on known physical relationships \citep{bonavita2024}. An example is vertical wind speed, which is linked to horizontal wind speed through the fundamental principle of continuity. However, as documented in previous studies \citep{selz2023,bonavita2024,hakim2024}, physical consistency is not always given.

In atmospheric science, developing physics-informed approaches, such as incorporating full partial differential equations (PDEs) like the Navier–Stokes equations directly into the loss function as done by \citet{raissi2019}, can be difficult. In practice, production-grade weather models like the Weather Research and Forecasting Model \cite{skamarock2021} solve not only the momentum equations (Navier–Stokes), but also the conservation equations for heat and moisture, all in three dimensions and over time. Consequently, a faithful PDE constraint in atmospheric physics would have to enforce this full, coupled 3D, time-dependent system. By contrast, the PDE-based constraints considered in the literature are often limited to idealized 2D flow cases (e.g., \citep{raissi2019,kochkov2021,fan2025}). 
Another complication is that these equations are typically solved on terrain-following native grids, whereas widely used reanalysis products such as ERA5 are provided on pressure levels after interpolation from the native model grid. This transformation can disrupt physical balances, making direct application of atmospheric PDEs as constraints unreliable if the interpolation effects are ignored. %The earlier discussion on gradient computation in complex terrain illustrates this issue.  
Furthermore, atmospheric flows are influenced by many processes beyond the resolved dynamics, including orography, surface properties (e.g., roughness, heat capacity), solar radiation, cloud formation, and precipitation. Many of these are parameterized in numerical models and enter the PDE solver as forcings or as auxiliary equations coupled to the governing dynamics. Applying the full set of atmospheric PDEs with all relevant forcings on the correct native grid would therefore require substantial additional development. A practical compromise is to incorporate physical principles into the loss function design, as we explore in this paper.

\paragraph{Spectral loss} Many statistical downscaling studies rely on deterministic mappings \citep{harilal2022enhancedsd, Kajbaf2022, gmd-18-161-2025, Hohlein2020WindDownscaling}, where the task reduces to estimating the conditional mean $\boldsymbol{\mu}=\mathbb{E}[\mathbf{q}|\mathbf{x}]$. Deterministic estimates of \(\mu\) are obtained by minimizing a standard pixel-space loss (\(\ell_1\) or \(\ell_2\) loss), which often produce blurry predictions~\citep{Bracco2025}. Recently, attention has shifted to probabilistic approaches \citep{merizzi2024wind, fotiadis2025adaptive, mardani2025} that aim to learn the full conditional distribution $p(\mathbf{q}|\mathbf{x})$, inspired by the remarkable success of diffusion models for super‑resolution tasks in computer vision. The probabilistic nature of these models allows them to quantify uncertainty and achieve lower pixel‑space errors (e.g., MAE) \citep{merizzi2024wind, fotiadis2025adaptive, mardani2025}. However, whether these predictions are also physically consistent remains an open question.

 Previous work has shown that incorporating Fourier-based losses helps restore high-frequency details in efficient super-resolution models \citep{fuoli2021fourier, yan2024fourier}. Our PSD-based loss motivation is rooted in the physics of weather and climate data: large-scale structures (e.g., pressure systems) and fine-scale details (e.g., local wind shifts, sharp temperature gradients) coexist, and the PSD quantifies how much variance is present at each spatial scale. If a downscaling model reproduces the large-scale features but underestimates smaller ones, its PSD will decay too rapidly at high frequencies, indicating missing small-scale physics. By incorporating a PSD-based loss, we explicitly encourage the model to match the observed scale-by-scale variance distribution, aiming to improve physical realism and mitigate both generalization and physics-consistency shortcomings.
\section{Methods}
The next subsections present physical consistency checks comparing key variables like kinetic energy, divergence, and vorticity, followed by a spectral loss term based on PSD to improve fine-scale detail. Together, they evaluate the physical realism and spatial accuracy of the model predictions.

\subsection{Physical consistency checks}\label{par:phys_checks} 

In this work, to evaluate the physical consistency of the model predictions, we employ variable interdependencies known from fluid mechanics and meteorology, similar to \citep{bonavita2024}.
More specifically, we compare each ground truth sample and its corresponding model prediction with respect to the following variables:
\begin{enumerate}
    \item Mean horizontal kinetic energy, defined as
    \begin{equation}
    \mathbf{E}_h = \frac{\mathbf{u}^2  + \mathbf{v}^2}{2}.
    \label{eq:kinetic_energy}
    \end{equation}
    \item Horizontal continuity/divergence, given by
    \begin{equation}
    \boldsymbol{\delta}_h = \nabla_h \cdot \mathbf{v}_h
    = \frac{\partial \mathbf{u}}{\partial x} + \frac{\partial \mathbf{v}}{\partial y}.
    \label{eq:divergence}
    \end{equation}
    \item Relative horizontal vorticity, expressed as
    \begin{equation}
    \boldsymbol{\zeta}_h = \nabla_h \times \mathbf{v}_h
    = \frac{\partial \mathbf{v}}{\partial x} - \frac{\partial \mathbf{u}}{\partial y}.
    \label{eq:vorticity}
    \end{equation}
\end{enumerate}

% These metrics can also be evaluated for a full 3D velocity field $\mathbf{v} = (\mathbf{u}, \mathbf{v}, \mathbf{w})$, but as our datasets only contain horizontal 2D fields ($\mathbf{v}_h = (\mathbf{u}, \mathbf{v})$), we utilize the horizontal versions.
Here, $\mathbf{u}$ and \(\mathbf{v}\) are the mean horizontal wind components in the zonal ($x$) and meridional ($y$) directions, respectively. Since statistical downscaling does not consider the temporal evolution of a meteorological state, more complex dynamical physics-consistency checks (see, e.g., \citep{selz2023,hakim2024}) are not applicable.
% We stress that these variables are computed offline, based solely on predictions of $\mathbf{u}$ and $\mathbf{v}$, so they do not enter the optimization.

Physically and mathematically, the three variables allow checking for different constraints that a realistic and physically consistent wind field must satisfy.  
Comparing $\mathbf{E}_h$ between ground truth and prediction, for example, assesses whether the nonlinear relationship between $\mathbf{u}$ and $\mathbf{v}$ is preserved.  
Checking if divergence $\boldsymbol{\delta}_h$ and vorticity $\boldsymbol{\zeta}_h$ match is crucial, as a realistic differentiable wind vector field can be decomposed into a divergence-free and a vorticity-free component via the Helmholtz decomposition \citep{dutton1986,bonavita2024}.  
%The divergence of a 3D flow field is also related to mass conservation, so requiring horizontal 2D divergence to agree between ground truth and prediction implicitly balances horizontal and vertical flow.  
Mathematically, the gradients in $\boldsymbol{\delta}_h$ and $\boldsymbol{\zeta}_h$ allow us to test whether the models capture the differential relationships between their outputs $\mathbf{u}$ and $\mathbf{v}$.
% Following a similar motivation, these expressions are used as losses in a previous study \citep{sekiyama2023}, but for fair assessment, we only utilize them as metrics.
% Disclaimer complex terrain

We note that, strictly speaking, all gradients should be horizontal gradients, which is not satisfied in complex terrain when $\mathbf{u}$ and $\mathbf{v}$ are at 10\,m height above the terrain (see Table~\ref{tab:variables}).  
Consequently, numerical values in mountainous regions are overestimated \cite{jacobson2005}, but the metrics remain useful for broadly assessing physical consistency.

% Note, however, that the horizontal gradients in our case are computed from $u$ and $v$ at 10\,m height above the terrain (cf.~Tab~\ref{tab:variables}).
% In complex terrain, such as mountains, neighboring grid points may have a significant vertical difference between them, so that the velocity gradients are not actually horizontal. % (i.e., parallel to the mean sea level) but follow the topography.
% While this leads to an overestimation of values in complex terrain compared to flat terrain \cite{jacobson2005}, the above considerations still hold, making the metrics valuable to assess the physical consistency of model predictions.\\

\subsection{PSD Loss Term} 
To encourage fidelity at high spatial frequencies, we include a PSD penalty alongside the pre-defined loss of the considered baselines: 
\(\mathcal{L}_{\text{total}} = \mathcal{L}_{\text{baseline}} + \lambda\,\mathcal{L}_{\text{PSD}}\). 
The specific baselines are described in Section~\ref{experimental-results}.

To elaborate, let \(\mathbf{q} \in \mathbb{R}^{H\times W}\) denote a single-channel atmospheric variable field on a grid of height \(H\) and width \(W\) 
(here \(\mathbb{R}^{H\times W}\) denotes the set of \(H\times W\) real matrices).  
For multi-channel data (e.g., multiple atmospheric variables), the PSD is computed independently for each channel.  
The PSD is obtained from the squared magnitude of the Fourier coefficients, normalized by the spatial dimensions and the grid spacing \(\Delta x\):
\begin{equation}
\mathrm{PSD}(\mathbf{q})(k_h,k_w) = \frac{\left|\mathcal{F}(\mathbf{q})(k_h,k_w)\right|^2}{H\,W\,\Delta x}\,,
\end{equation}
where \(k_h = 0,\dots,H-1\) and \(k_w = 0,\dots,W-1\) are the discrete wavenumbers in the vertical and horizontal directions, respectively, and \(\mathcal{F}(\mathbf{q})\) is the 2D discrete Fourier transform.  

To define $\mathcal{L}_{\text{PSD}}$, we weigh the contribution of each PSD coefficient in proportion to its normalized isotropic wavenumber squared. The isotropic wavenumber $k$ is obtained from \(k_h\) and \(k_w\) via
\(k = \sqrt{k_h^2 + k_w^2}\), and the weights are computed as \(w(k_h,k_w)=(k / k_{\max})^2\). This quadratic weighting emphasizes high-frequency components, where the largest discrepancies occur, and therefore guides the model to reduce over-smoothing at small spatial scales.

The weighted PSD loss between ground truth \(\mathbf{q}\) and prediction \(\hat{\mathbf{q}}\) is then calculated as:
\begin{equation}
\mathcal{L}_{\text{PSD}} =
\sqrt{\frac{1}{H\,W}\sum_{k_h=0}^{H-1}\sum_{k_w=0}^{W-1}
w(k_h,k_w)\,
\left[
\log\bigl(\mathrm{PSD}(\mathbf{q})(k_h,k_w)\bigr)
-
\log\bigl(\mathrm{PSD}(\hat{\mathbf{q}})(k_h,k_w)\bigr)
\right]^2}.
\end{equation}
%where \(w(k_h,k_w) = \left(k / k_{\max}\right)^2\) is the high-frequency weighting.  
%In later plots and discussion, \(k\) denotes this isotropic wavenumber.

%%%%%%%%%%% PLOTTED results - U wind - Central Europe
\begin{figure}[htbp!]
  \centering
  % Row 1: three equally spaced images (a, b, c)
  \begin{subfigure}[t]{0.26\textwidth}
    \centering
    \includegraphics[width=\linewidth]{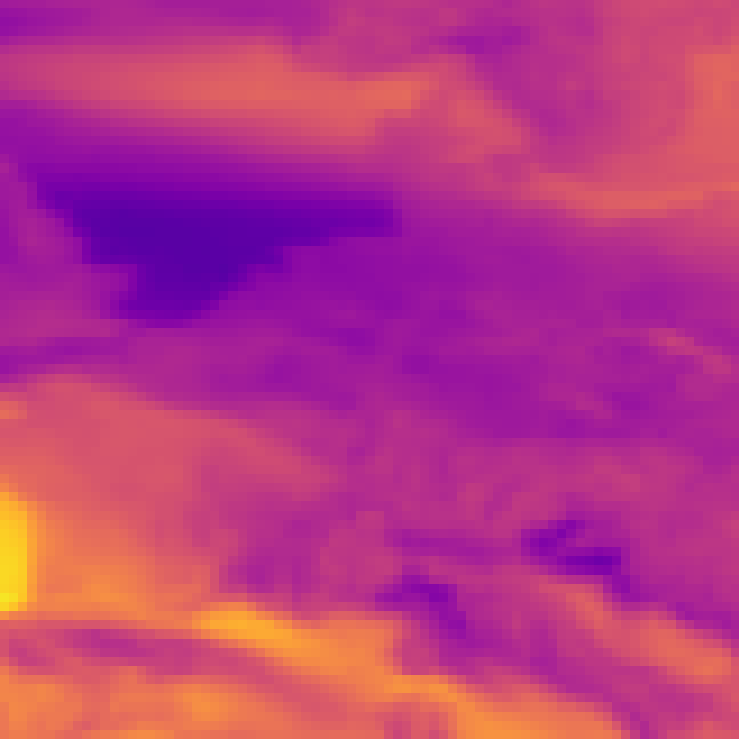}
    \subcaption{Input: ERA5}
    
  \end{subfigure}
  \hfill
  \begin{subfigure}[t]{0.26\textwidth}
    \centering
    \includegraphics[width=\linewidth]{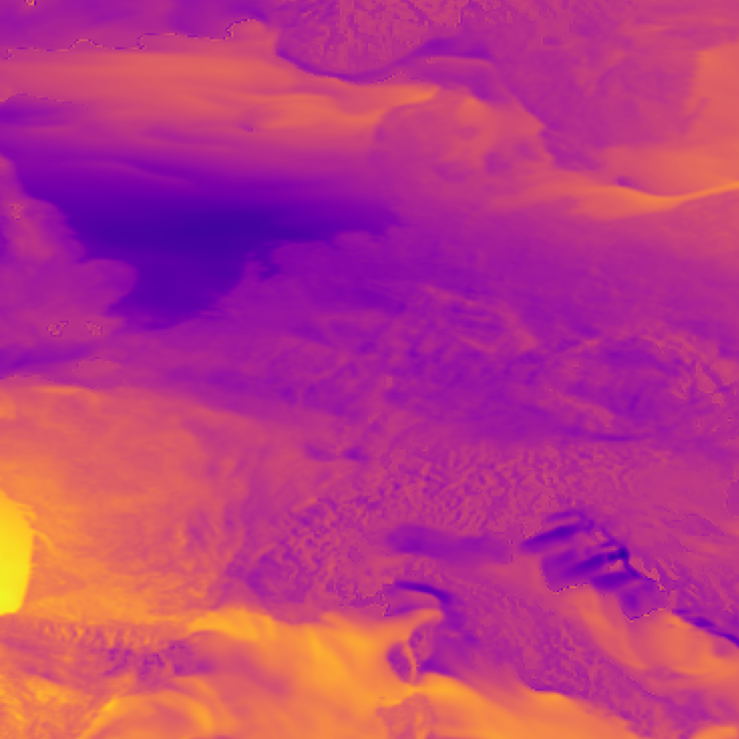}
    \subcaption{Target: CERRA}
    
  \end{subfigure}
  \hfill
  \begin{subfigure}[t]{0.26\textwidth}
    \centering
    \includegraphics[width=\linewidth]{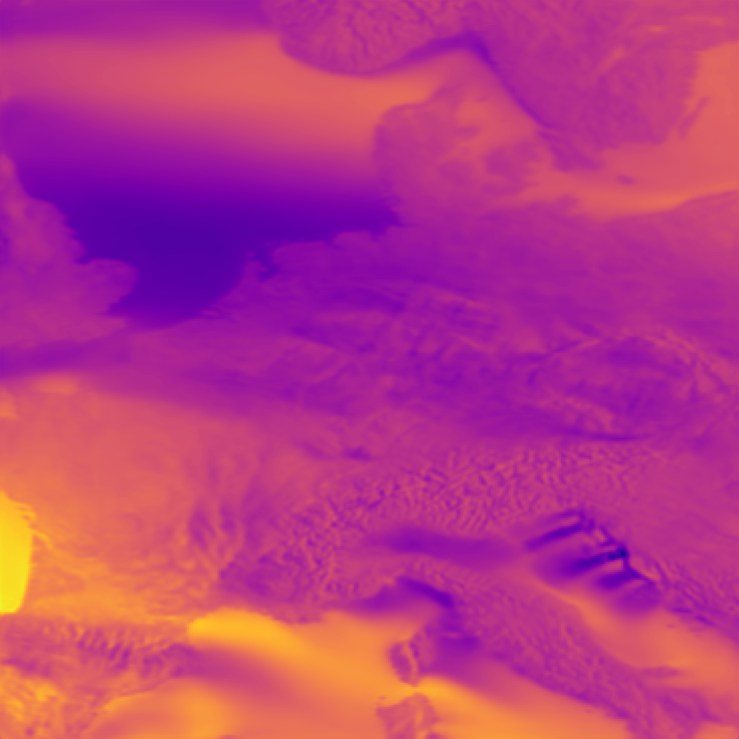}
    \subcaption{CRPS UNets-PSD}
    
  \end{subfigure}

  \vspace{0.7em}

  % Row 2: full-width colormap
  \begin{subfigure}[t]{0.5\textwidth}
    \centering
    \includegraphics[width=\linewidth]{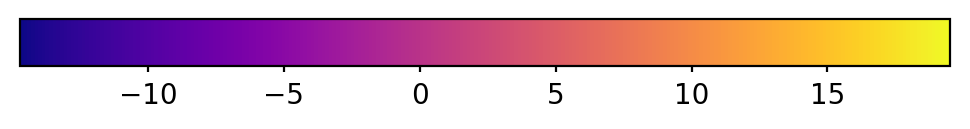}
    
  \end{subfigure}
   \vspace{-0.8em} %
  \caption{Input, target, and CRPS UNets-PSD prediction of $\mathbf{u}$ in Central Europe (in-distribution region) on 1st of January 2021}
  \label{fig:preds-CE-maintext}
\end{figure}

%%%%%%%%%%% PLOTTED results - U wind - Iberia
\begin{figure}[htbp!]
  \centering
  % Row 1: three equally spaced images (a, b, c)
  \begin{subfigure}[t]{0.26\textwidth}
    \centering
    \includegraphics[width=\linewidth]{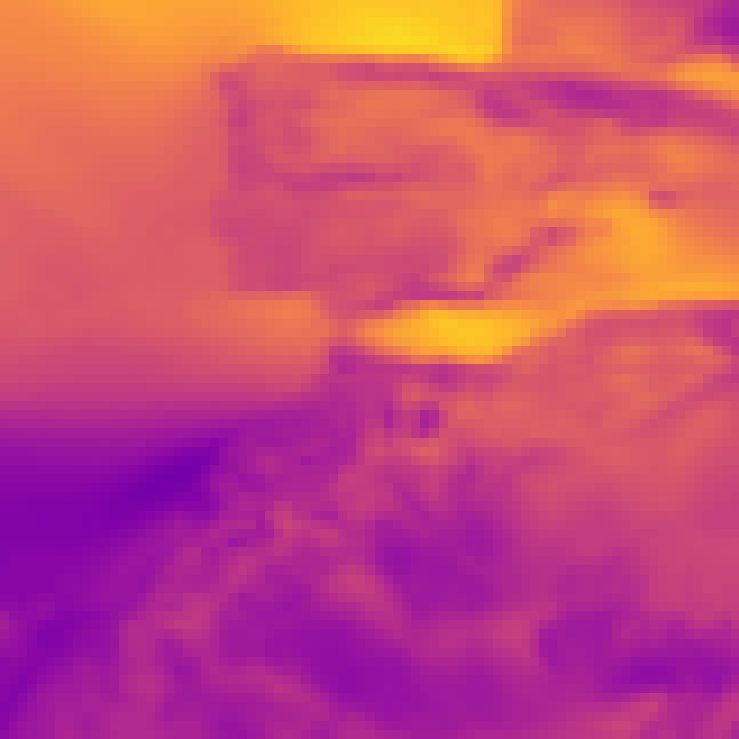}
    \subcaption{Input: ERA5}
    
  \end{subfigure}
  \hfill
  \begin{subfigure}[t]{0.26\textwidth}
    \centering
    \includegraphics[width=\linewidth]{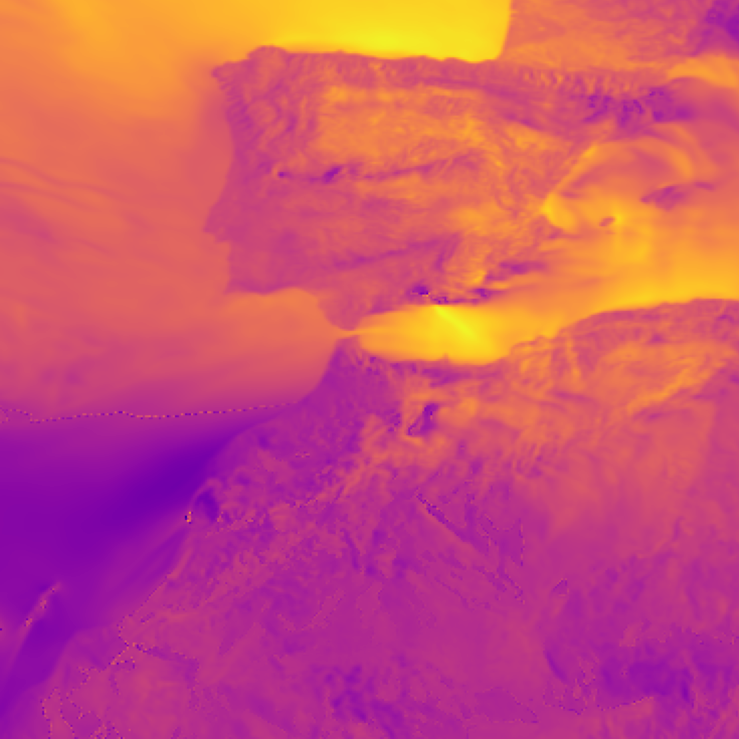}
    \subcaption{Target: CERRA}
    
  \end{subfigure}
  \hfill
  \begin{subfigure}[t]{0.26\textwidth}
    \centering
    \includegraphics[width=\linewidth]{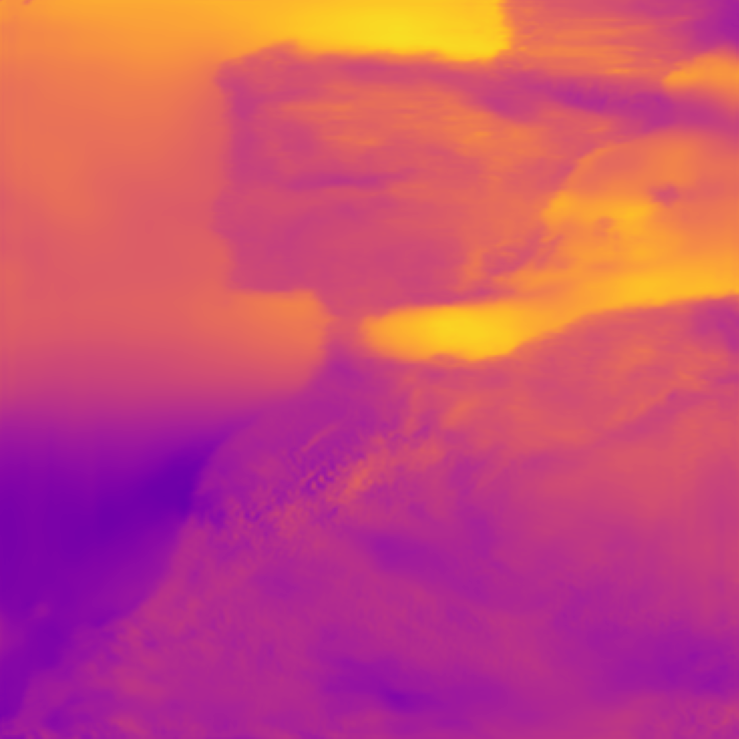}
    \subcaption{CRPS UNets-PSD}
    
  \end{subfigure}

  \vspace{0.7em}

  % Row 2: full-width colormap
  \begin{subfigure}[t]{0.5\textwidth}
    \centering
    \includegraphics[width=\linewidth]{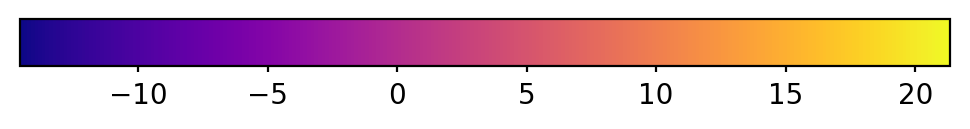}
    
  \end{subfigure}
   \vspace{-0.8em}
  \caption{Input, target, and CRPS UNets-PSD prediction of $\mathbf{u}$ in the Iberia, Morocco region (OOD) on 1st of January 2021. The prediction is much blurrier compared to in-distribution as shown Figure~\ref{fig:preds-CE-maintext}. }
  \label{fig:preds-Iberia-maintext}
\end{figure}

%------------------------------------------------------------------------------------------------------------------------------
%%%%% Experiments setup %%%%%%%%%%%
%------------------------------------------------------------------------------------------------------------------------------
\vspace{-0.8em}
\section{Experimental results}
\subsection{Experiments setup}
\label{experimental-results}
We benchmark three state-of-the-art downscalers: the full probabilistic \textit{CorrDiff}  model of \citet{mardani2025}, its deterministic regression backbone (\textit{Regression-CorrDiff}), and a probabilistic U-Net ensemble trained with a continuous ranked probability score (CRPS) loss (\textit{CRPS-UNets}) following \citet{alet2025skillfuljointprobabilisticweather}. Each architecture is retrained from scratch in two variants, one using its standard loss and one augmented with the extra PSD loss term. All models take as input the set of coarse-resolution variables listed in Table~\ref{tab:variables}, taken from the ERA5 reanalysis at 25\,km resolution \citep{era5} on multiple pressure levels. Their task is to predict the corresponding high-resolution surface fields from the CERRA reanalysis at 5.5\,km resolution \citep{cerra}: 10\,m U-component of wind ($\mathbf{u}$), 10\,m V-component of wind ($\mathbf{v}$), and 2\,m temperature ($\mathbf{t2m}$).  
Training is performed on an ERA5–CERRA subset cropped to Central Europe. To test geographic generalization, we create two additional, non-overlapping ERA5–CERRA subsets, one over the Iberia, Morocco region and one over Northern Scandinavia. They are used only for OOD evaluation and not for training or fine-tuning. We assess physical consistency as indicated in Section~\ref{par:phys_checks}. Further information on the datasets and training details can be found in Appendix~\ref{section:Dataset details}.

\begin{table}[htbp!]
\centering
\caption{Meteorological variables used as inputs and outputs for the downscaling models.}
\begin{tabular}{ll}
\toprule
\textbf{Input variables} & \textbf{Output variables} \\ 
\midrule
\multicolumn{2}{l}{\emph{Pressure levels (500\,hPa and 850\,hPa)}} \\
\quad Temperature                 &  \\[-2pt]
\quad U-wind component            &  \\[-2pt]
\quad V-wind component            &  \\[4pt]
\multicolumn{2}{l}{\emph{Surface level}} \\
\quad 10\,m U-component of wind   & 10\,m U-component of wind \\[-2pt]
\quad 10\,m V-component of wind   & 10\,m V-component of wind \\[-2pt]
\quad 2\,m Temperature            & 2\,m Temperature \\[-2pt]
\bottomrule
\end{tabular}
\label{tab:variables}
\end{table}

\subsection{Results} 
\begin{figure}[htbp!]
  \centering
  %------------------ 1st ROW ------------------
  \begin{subfigure}[t]{0.33\textwidth}
    \centering
    \includegraphics[width=\linewidth]{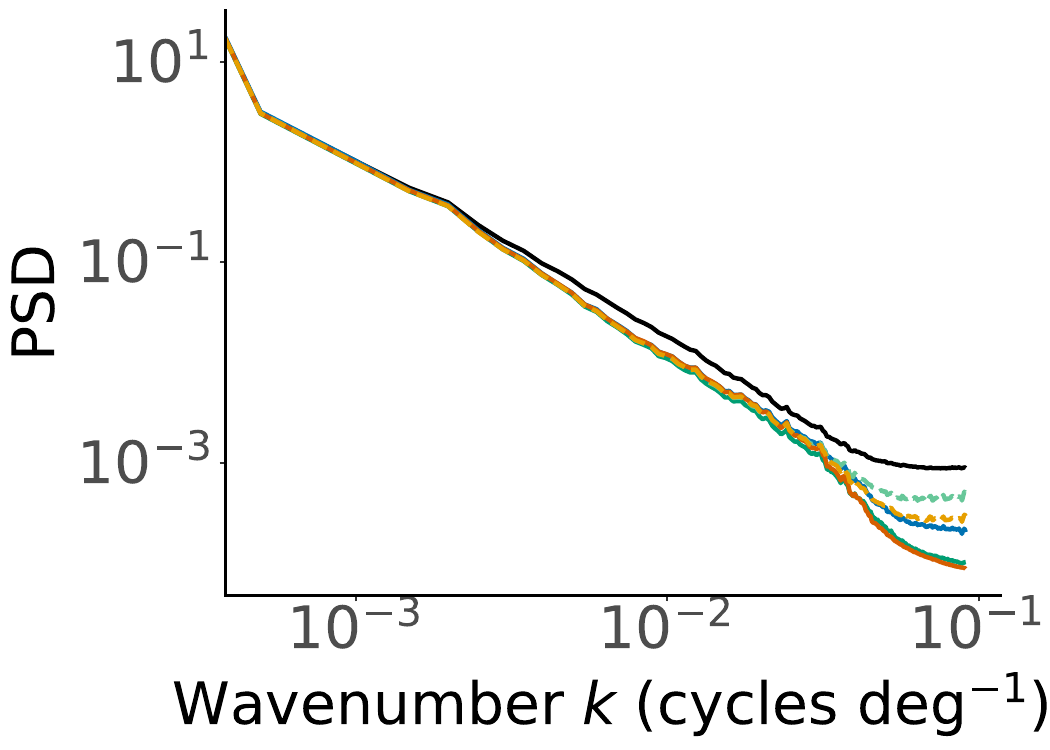}
    \caption{PSD of $\mathbf{u}$}
  \end{subfigure}\hfill
  \begin{subfigure}[t]{0.33\textwidth}
    \centering
    \includegraphics[width=\linewidth]{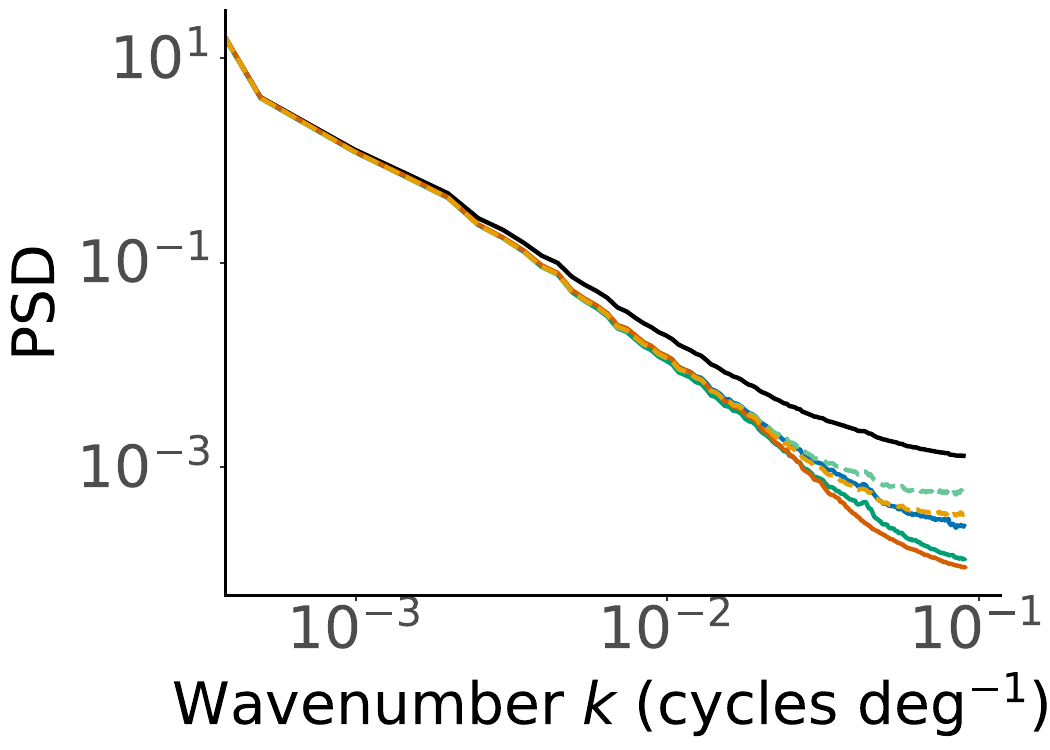}
    \caption{PSD of $\mathbf{v}$}
  \end{subfigure}\hfill
  \begin{subfigure}[t]{0.33\textwidth}
    \centering
    \includegraphics[width=\linewidth]{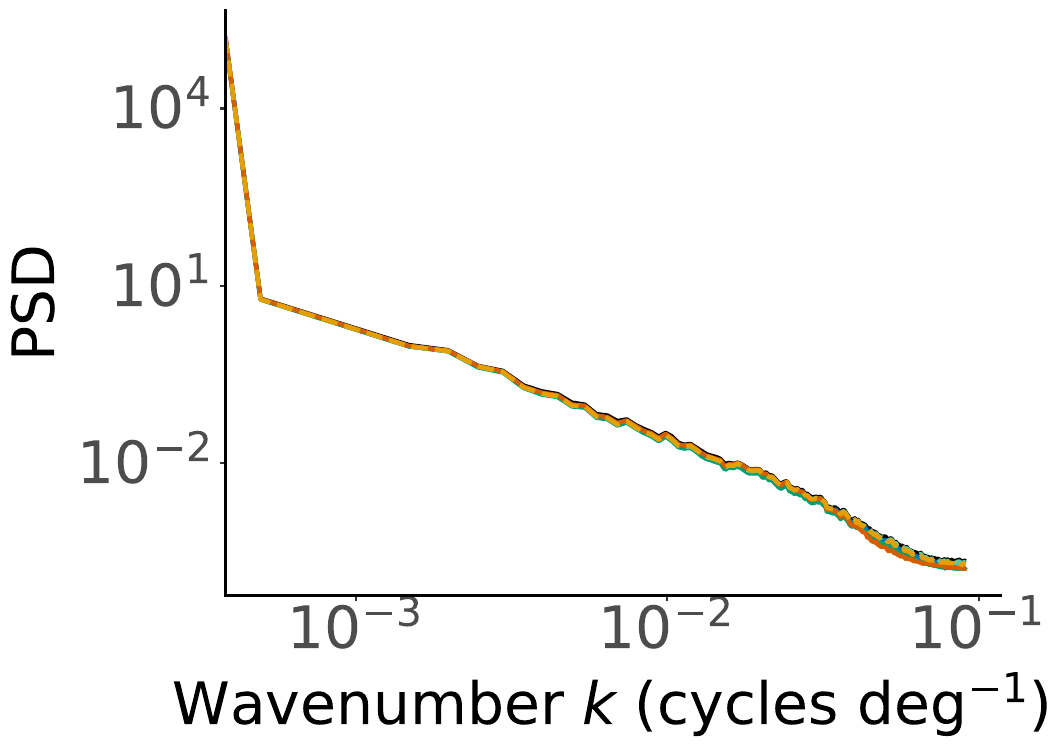}
    \caption{PSD of temperature $\mathbf{t2m}$}
  \end{subfigure}

  \vspace{0.5em} % vertical gap between rows
  
  %------------------ 2nd ROW ------------------
  \begin{subfigure}[t]{0.33\textwidth}
    \centering
    \includegraphics[width=\linewidth]{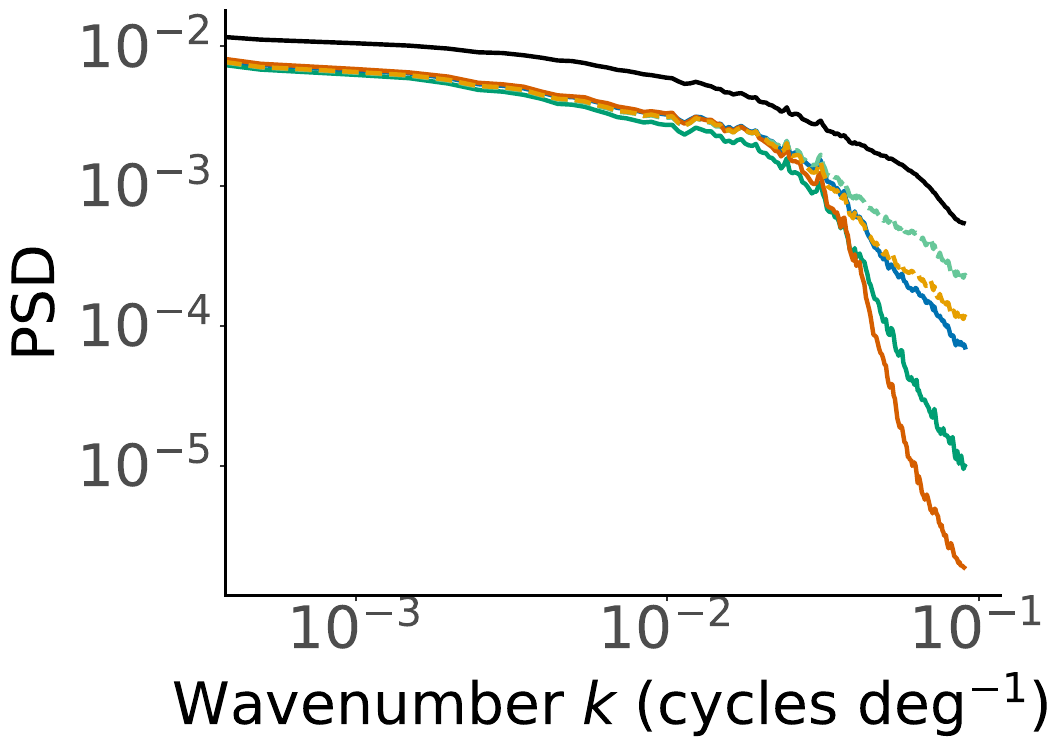}
    \caption{PSD of $\boldsymbol{\delta_h}$}
  \end{subfigure}\hfill
  \begin{subfigure}[t]{0.33\textwidth}
    \centering
    \includegraphics[width=\linewidth]{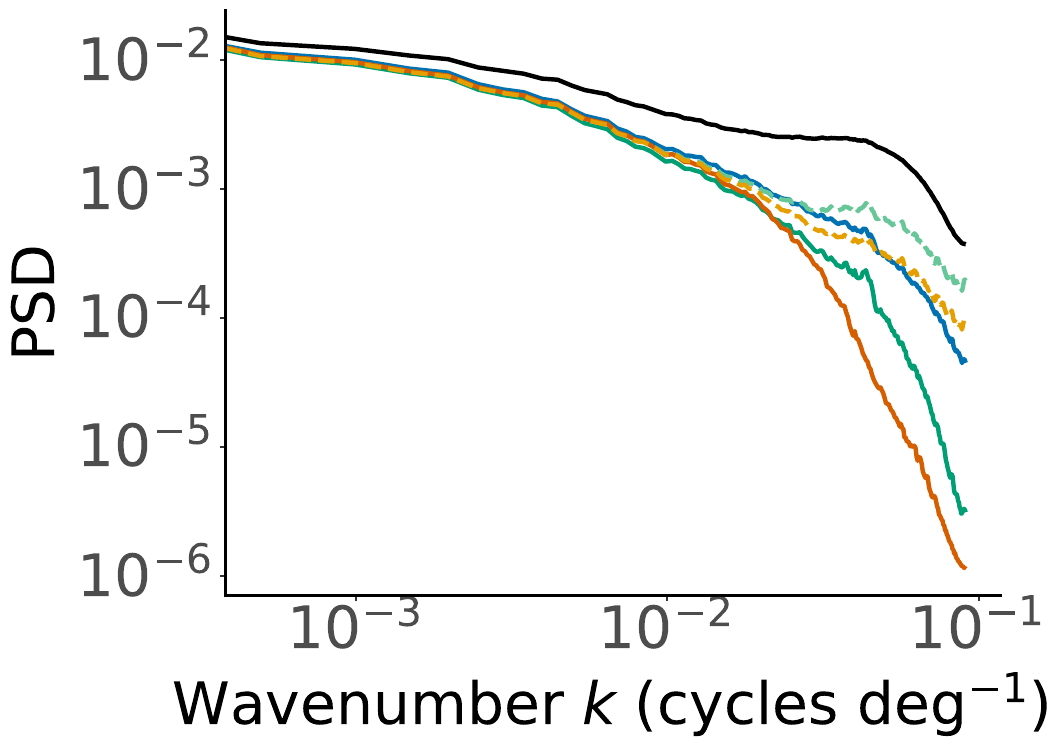}
    \caption{PSD of $\boldsymbol{\zeta}_h$}
  \end{subfigure}\hfill
  \begin{subfigure}[t]{0.33\textwidth}
    \centering
    \includegraphics[width=\linewidth]{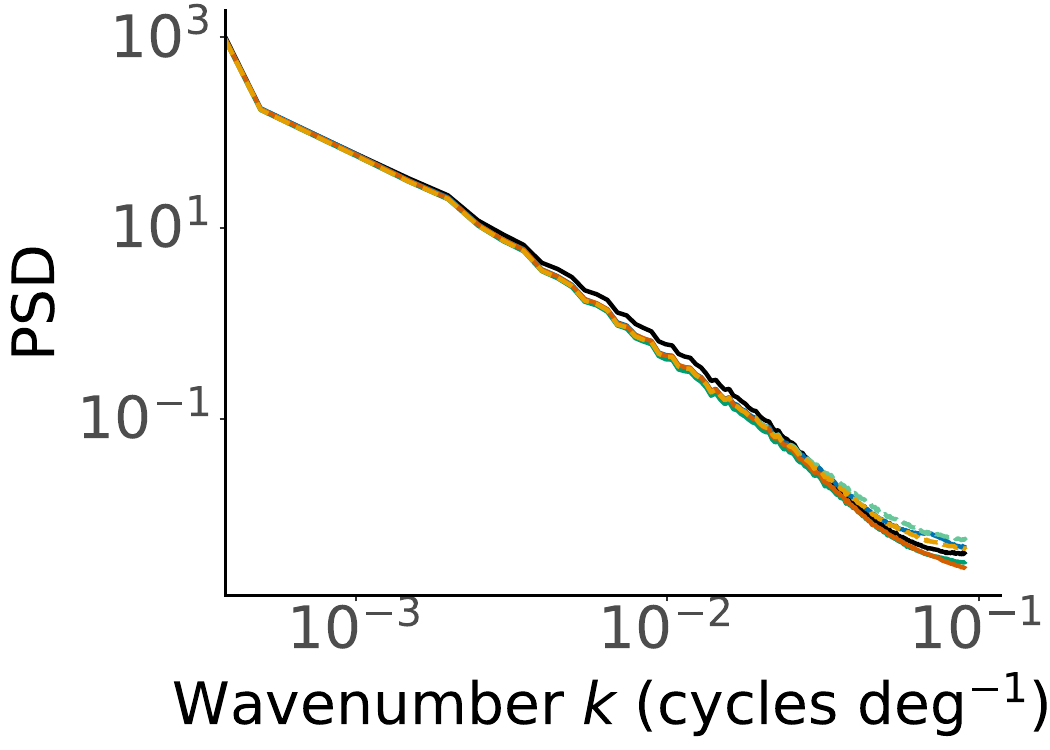}
    \caption{PSD of $\mathbf{E}_h$}
  \end{subfigure}

  \caption{PSD comparison of down-scaled (top row) and physics-derived (bottom row) variables from CERRA (ground truth) and model predictions in Central Europe (in-distribution domain). Wavenumber \(k\) is plotted on a logarithmic scale.}
  \label{fig:psd_ce_all}
\end{figure}

Table~\ref{tab:all_regions_mae_rmse} shows the mean-absolute error (MAE) and root-mean-square error (RMSE) for different models over Central Europe, the Iberia, Morocco region, and Northern Scandinavia, while Table~\ref{tab:all_regions_crps} reports the CRPS. Results for \textit{Regression-CorrDiff} and \textit{Regression-CorrDiff-PSD} are omitted from this latter table as these models are deterministic, for which CRPS is equivalent to MAE \citep{hersbach2000decomposition}. Figure~\ref{fig:psd_ce_all} shows the comparison of PSD curves computed between CERRA (ground truth) and model predictions.

Table~\ref{tab:all_regions_mae_rmse} indicates that, in the in-distribution test over Central Europe, the three baseline models produce accurate downscalings showing low MAE and RMSE. Table~\ref{tab:all_regions_crps} further shows that CRPS values for \textit{CorrDiff} and \textit{CRPS UNets} are also low, consistent with the results reported by \citet{mardani2025}. From Figure~\ref{fig:psd_ce_all}, we note that among the target variables, \(\mathbf{t2m}\) is the most well-behaved and easy to predict: it exhibits the lowest errors and shows little sensitivity to the PSD loss, reflecting its smoother spatial structure and weaker small-scale variability compared to wind components \citep{gmd-18-2051-2025}. 
The PSDs of $\mathbf{u}$, $\mathbf{v}$, and \(\mathbf{E}_h\) closely follow the reference spectrum, whereas those of \(\boldsymbol{\zeta}_h\) and \(\boldsymbol{\delta}_h\) deviate markedly. Particularly, the second row of Figure~\ref{fig:psd_ce_all} indicates that these predictions are not fully physically consistent. This gap narrows when both \textit{Regression-CorrDiff} and  \textit{CRPS-UNets} are trained with the PSD loss term, returning spectral curves that follow the reference more closely. Notably, after adding the PSD loss, \textit{Regression-CorrDiff} surpasses the full probabilistic \textit{CorrDiff} architecture in spectral fidelity, suggesting that frequency-domain regularization can offset some of the advantages typically associated with probabilistic approaches. 
Training \textit{CorrDiff}'s diffusion component on the residuals of the \textit{Regression-CorrDiff-PSD} backbone yields no noticeable improvement in our experiments. We hypothesize that training \textit{Regression-CorrDiff} with a PSD loss already captures high-frequency content, leaving little signal for the diffusion stage. Consequently, the diffusion step adds negligible skill and we therefore omit that variant.

\begin{figure}[htbp!]
  \centering
  \includegraphics[width=0.7\linewidth]{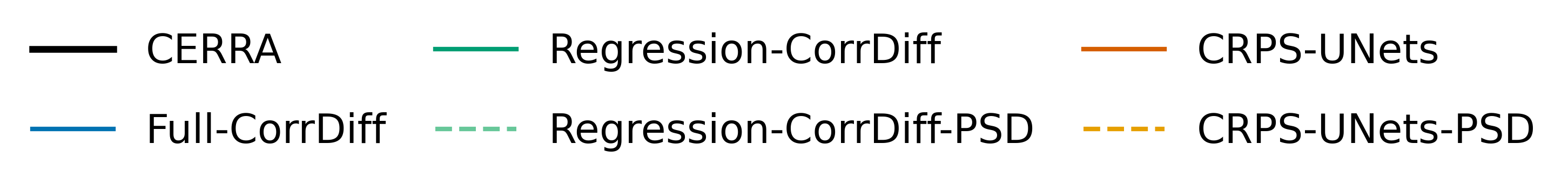}
\end{figure}
\vspace{-8pt}
%%%%%%%%%%% PSD CENTRAL EUROPE – ALL VARIABLES IN ONE FIGURE

In OOD domains (the Iberia, Morocco region and Northern Scandinavia), MAE and RMSE rise sharply relative to the in-distribution case (Table~\ref{tab:all_regions_mae_rmse}), and CRPS values (Table~\ref{tab:all_regions_crps}) also increase. Models trained with the PSD loss term achieve slightly better scores, yet, their performance remains well below the in-distribution level and predictions look blurrier (see Figure~\ref{fig:preds-CE-maintext} and~\ref{fig:preds-Iberia-maintext}). For more extensive visual results see Appendix~\ref{app:results-CE},~\ref{app:results-Iberia}, and~\ref{app:results-Scandinavia}.

We also observe that variables the model is directly optimized for, such as \(\mathbf{u}\) and \(\mathbf{v}\), tend to match the ground truth more closely, particularly at low frequencies. This is expected, as loss functions like MSE emphasize minimizing errors in low-frequency components and tend to suppress high-frequency features such as sharp gradients. In contrast, derived variables such as \(\boldsymbol{\delta}_h\) and \(\boldsymbol{\zeta}_h\) exhibit a consistent offset from the ground truth, even at low frequencies. This suggests that the model is not merely missing high-frequency details, but is also failing to capture important underlying correlations between physical variables: the model can learn to reproduce wind components while missing the physics embedded in their spatial derivatives.

%%%%%%%%%%% MAE, RMSE — All Regions %%%%%%%%%%%
\begin{table}[htbp!]
\centering
\caption{MAE and RMSE of $\mathbf{u}$, $\mathbf{v}$, and $\mathbf{t2m}$ over Central Europe (in-distribution) and the Iberia, Morocco region / Northern Scandinavia (OOD).}
\scriptsize
\setlength{\tabcolsep}{4pt}
\renewcommand{\arraystretch}{1.1}
\begin{tabular}{llrrrrrrrrrr}
\toprule
Region & Variable
& \multicolumn{2}{c}{CorrDiff}
& \multicolumn{2}{c}{Reg-CorrDiff}
& \multicolumn{2}{c}{Reg-CorrDiff-PSD}
& \multicolumn{2}{c}{CRPS UNets}
& \multicolumn{2}{c}{CRPS UNets-PSD} \\
& &
MAE & RMSE & MAE & RMSE & MAE & RMSE & MAE & RMSE & MAE & RMSE \\
\midrule
\multirow{3}{*}{Central Europe}
  & $\mathbf{u}$, m/s   & \textbf{0.663} & \textbf{0.971} & 0.674 & 0.978 & 0.676 & 0.987 & 0.671 & 0.977 & 0.678 & 0.982 \\
  & $\mathbf{v}$, m/s   & \textbf{0.669} & \textbf{1.014} & 0.681 & 1.026 & 0.685 & 1.040 & 0.679 & 1.037 & 0.690 & 1.037 \\
  & $\mathbf{t2m}$, K   & \textbf{0.602} & \textbf{0.826} & 0.622 & 0.852 & 0.619 & 0.857 & 0.619 & 0.855 & 0.617 & 0.853 \\
\midrule
\multirow{3}{*}{Iberia, Morocco}
  & $\mathbf{u}$, m/s   & 1.067 & 1.446 & 1.074 & 1.459 & 1.063 & 1.446 & 1.044 & 1.407 & \textbf{1.035} & \textbf{1.398} \\
  & $\mathbf{v}$, m/s   & 1.094 & 1.528 & 1.080 & 1.523 & 1.071 & 1.517 & 1.044 & 1.469 & \textbf{1.042} & \textbf{1.460} \\
  & $\mathbf{t2m}$, K   & 1.486 & 2.118 & 1.459 & 2.088 & 1.426 & 2.062 & 1.361 & 1.937 & \textbf{1.323} & \textbf{1.857} \\
\midrule
\multirow{3}{*}{Northern Scandinavia}
  & $\mathbf{u}$, m/s   & 1.021 & 1.355 & 1.042 & 1.398 & 0.999 & 1.332 & 0.932 & 1.258 & \textbf{0.858} & \textbf{1.156} \\
  & $\mathbf{v}$, m/s   & 1.001 & 1.338 & 1.010 & 1.365 & 0.981 & 1.344 & 0.925 & 1.225 & \textbf{0.875} & \textbf{1.174} \\
  & $\mathbf{t2m}$, K   & 1.481 & 1.883 & 1.471 & 1.908 & 1.390 & 1.823 & \textbf{1.304} & \textbf{1.677} & 1.405 & 1.835 \\
\bottomrule
\end{tabular}
\label{tab:all_regions_mae_rmse}
\end{table}

\begin{table}[htbp!]
\centering
\caption{CRPS of $\mathbf{u}$, $\mathbf{v}$, and $\mathbf{t2m}$ over Central Europe (in-distribution) and the Iberia, Morocco region / Northern Scandinavia (OOD). }
\scriptsize
\setlength{\tabcolsep}{6pt}
\renewcommand{\arraystretch}{1.1}
\begin{tabular}{llrrr}
\toprule
Region & Variable
& CorrDiff
& CRPS UNets
& CRPS UNets-PSD \\
\midrule
\multirow{3}{*}{Central Europe}
  & $\mathbf{u}$, m/s   & \textbf{0.480} & 0.515 & 0.516 \\
  & $\mathbf{v}$, m/s   & \textbf{0.476} & 0.520 & 0.520 \\
  & $\mathbf{t2m}$, K   & \textbf{0.432} & 0.470 & 0.467 \\
\midrule
\multirow{3}{*}{Iberia, Morocco}
  & $\mathbf{u}$, m/s   & 0.828 & 0.801 & \textbf{0.786} \\
  & $\mathbf{v}$, m/s   & 0.849 & 0.815 & \textbf{0.797} \\
  & $\mathbf{t2m}$, K   & 1.197 & 1.077 & \textbf{1.033} \\
\midrule
\multirow{3}{*}{Northern Scandinavia}
  & $\mathbf{u}$, m/s   & 0.760 & 0.718 & \textbf{0.661} \\
  & $\mathbf{v}$, m/s   & 0.750 & 0.727 & \textbf{0.692} \\
  & $\mathbf{t2m}$, K   & 1.102 & 1.017 & \textbf{1.075} \\
\bottomrule
\end{tabular}
\label{tab:all_regions_crps}
\end{table}

%------------------------------------------------------------------------------------------------------------------------------
%%%%% Discussion %%%%%%%%%%%
%------------------------------------------------------------------------------------------------------------------------------
\subsection{Qualitative discussion}
In atmospheric dynamics, preserving the PSD slope across different scales is crucial: deviations imply that the model redistributes variance incorrectly among scales, leading to either excessive smoothing (loss of small-scale structure) or spurious noise. Standard pixel-space losses (\(\ell_1\) or \(\ell_2\) loss) implicitly weight low-\(k\) modes more heavily, since large-scale errors dominate the sum of squared differences. This biases deep learning downscalers toward fitting coarse features accurately at the expense of mesoscale and sub-mesoscale dynamics.

Further, \(\boldsymbol{\delta}_h\) and \(\boldsymbol{\zeta}_h\) are derived from spatial derivatives of the prognostic variables \(\mathbf{u}\) and \(\mathbf{v}\).
In Fourier space, differentiation multiplies each mode by its wavenumber magnitude (see Appendix~\ref{app:fourier_derivative}), so derivative-based diagnostics amplify high-frequency components.
Thus, if the high-wavenumber part of the PSD is under-represented, these derived variables systematically lose variance, even if the base fields \(\mathbf{u}\) and \(\mathbf{v}\) have satisfactory accuracy in pixel-space metrics. This explains why our baselines reproduce \(\mathbf{E}_h\) more faithfully than \(\boldsymbol{\delta}_h\) or \(\boldsymbol{\zeta}_h\): \(\mathbf{E}_h\) depends on the square of the velocity itself (low- and mid-\(k\) dominated), whereas divergence and vorticity are explicitly high-\(k\) weighted.

By directly optimizing a PSD-based term, we reward matching the observed scale-by-scale variance distribution, rather than only minimizing pixel-space errors. This frequency-domain regularization reduces the high-\(k\) deficit in \(\mathbf{u}\) and \(\mathbf{v}\) and, through the \(k\)-weighting of derivatives, improves the physical consistency of \(\boldsymbol{\delta}_h\) and \(\boldsymbol{\zeta}_h\) as well.
\vspace{-5pt}
\section{Conclusion}
\vspace{-5pt}
Our findings confirm that the considered baselines struggle to generalize to unseen geographic regions, underscoring the need for downscaling models explicitly designed for both spatial transferability and physical fidelity. We have also shown that a simple spectral loss has a clear and consistent impact on the physical realism of the predictions: it substantially improves the match between predicted and reference spectra, particularly at high spatial frequencies, and narrows the gap for physically derived variables such as \(\boldsymbol{\delta}_h\) and \(\boldsymbol{\zeta}_h\). These results indicate that, while geographic transferability remains challenging, explicitly encouraging the reconstruction of scale-by-scale variance is an effective way to preserve small-scale physical structures. In future work, it would be interesting to incorporate kinetic energy, divergence, and vorticity as either soft constraints in the loss function to guide the optimization of the velocities, following~\citet{sekiyama2023} or integrating the constraints as part of the model architecture akin to other research in the fluid dynamics and machine learning literature~\cite{Fang02102020,Borde01022022,borde2023aerothermodynamicsimulatorsrocketdesign}.

\section*{Acknowledgments}
This publication is part of the project DAILSCOM: Atmospheric Turbulence Informed Machine Learning for Laser Satellite Communications with file number 20617 of the research programme Open Technology Program 2023-2, which is (partly) financed by the Dutch Research Council (NWO).

This work is also part of the project Delft AI4WF: Delft Artificial Intelligence for Weather Forecast with file number 2024.023 of the research programme Computing time of national computer systems, which is (partly) financed by the NWO under the grant \url{https://doi.org/10.61686/VYGRS56933}.

\newpage

%------------------------------------------------------------------------------------------------------------------------------
%%%%% References %%%%%%%%%%%
%------------------------------------------------------------------------------------------------------------------------------
\bibliographystyle{plainnat}
\bibliography{sample}

\begin{thebibliography}{43}
\providecommand{\natexlab}[1]{#1}
\providecommand{\url}[1]{\texttt{#1}}
\expandafter\ifx\csname urlstyle\endcsname\relax
  \providecommand{\doi}[1]{doi: #1}\else
  \providecommand{\doi}{doi: \begingroup \urlstyle{rm}\Url}\fi

\bibitem[Alet et~al.(2025)Alet, Price, El-Kadi, Masters, Markou, Andersson, Stott, Lam, Willson, Sanchez-Gonzalez, and Battaglia]{alet2025skillfuljointprobabilisticweather}
Ferran Alet, Ilan Price, Andrew El-Kadi, Dominic Masters, Stratis Markou, Tom~R. Andersson, Jacklynn Stott, Remi Lam, Matthew Willson, Alvaro Sanchez-Gonzalez, and Peter Battaglia.
\newblock Skillful joint probabilistic weather forecasting from marginals, 2025.
\newblock URL \url{https://arxiv.org/abs/2506.10772}.

\bibitem[Bader et~al.(2009)]{Bader2009ClimateModels}
David~C. Bader et~al.
\newblock Climate models: An assessment of strengths and limitations.
\newblock U.S. Climate Change Science Program Synthesis and Assessment Product 3.1, U.S. Department of Energy, Washington, D.C., 2009.
\newblock U.S. Climate Change Science Program Synthesis and Assessment Product 3.1.

\bibitem[Bi et~al.(2023)Bi, Xie, Zhang, Chen, Gu, and Tian]{Bi2023}
Kaifeng Bi, Lingxi Xie, Hengheng Zhang, Xin Chen, Xiaotao Gu, and Qi~Tian.
\newblock Accurate medium-range global weather forecasting with {3D} neural networks.
\newblock \emph{Nature}, 619\penalty0 (7970):\penalty0 533--538, 2023.

\bibitem[Bodnar et~al.(2025)Bodnar, Bruinsma, Lucic, Stanley, Allen, Brandstetter, Garvan, Riechert, Weyn, Dong, et~al.]{Bodnar2025}
Cristian Bodnar, Wessel~P Bruinsma, Ana Lucic, Megan Stanley, Anna Allen, Johannes Brandstetter, Patrick Garvan, Maik Riechert, Jonathan~A Weyn, Haiyu Dong, et~al.
\newblock A foundation model for the {Earth} system.
\newblock \emph{Nature}, pages 1--8, 2025.

\bibitem[Bonavita(2024)]{bonavita2024}
Massimo Bonavita.
\newblock On some limitations of current machine learning weather prediction models.
\newblock \emph{Geophysical Research Letters}, 51\penalty0 (12):\penalty0 e2023GL107377, 2024.
\newblock ISSN 1944-8007.
\newblock \doi{10.1029/2023GL107377}.

\bibitem[Bracco et~al.(2025)Bracco, Brajard, Dijkstra, Hassanzadeh, Lessig, and Monteleoni]{Bracco2025}
Annalisa Bracco, Julien Brajard, Henk~A Dijkstra, Pedram Hassanzadeh, Christian Lessig, and Claire Monteleoni.
\newblock Machine learning for the physics of climate.
\newblock \emph{Nature Reviews Physics}, 7\penalty0 (1):\penalty0 6--20, 2025.

\bibitem[de~Ocáriz~Borde et~al.(2022)de~Ocáriz~Borde, Sondak, and Protopapas]{Borde01022022}
Haitz~Sáez de~Ocáriz~Borde, David Sondak, and Pavlos Protopapas.
\newblock Convolutional neural network models and interpretability for the anisotropic reynolds stress tensor in turbulent one-dimensional flows.
\newblock \emph{Journal of Turbulence}, 23\penalty0 (1-2):\penalty0 1--28, 2022.
\newblock \doi{10.1080/14685248.2021.1999459}.
\newblock URL \url{https://doi.org/10.1080/14685248.2021.1999459}.

\bibitem[de~Ocáriz~Borde et~al.(2023)de~Ocáriz~Borde, Innocenzi, and Savarino]{borde2023aerothermodynamicsimulatorsrocketdesign}
Haitz~Sáez de~Ocáriz~Borde, Pietro Innocenzi, and Flavio Savarino.
\newblock Aerothermodynamic simulators for rocket design using neural fields, 2023.
\newblock URL \url{https://arxiv.org/abs/2303.10283}.

\bibitem[Dutton(1986)]{dutton1986}
John~A. Dutton.
\newblock \emph{The Ceaseless Wind: An Introduction to the Theory of Atmospheric Motion}.
\newblock Dover Publications, Inc., USA, 2 edition, 1986.
\newblock ISBN 0-486-65096-0.

\bibitem[Fallah et~al.(2025)Fallah, Rostami, Russo, Harder, Menz, Hoffmann, Didovets, and Hattermann]{gmd-18-161-2025}
Bijan Fallah, Masoud Rostami, Emmanuele Russo, Paula Harder, Christoph Menz, Peter Hoffmann, Iulii Didovets, and Fred~F Hattermann.
\newblock Climate model downscaling in central {{Asia}}: a dynamical and a neural network approach.
\newblock \emph{Geoscientific Model Development}, 18\penalty0 (1):\penalty0 161--180, 2025.

\bibitem[Fan et~al.()Fan, Akhare, and Wang]{fan2025}
Xiantao Fan, Deepak Akhare, and Jian-Xun Wang.
\newblock Neural differentiable modeling with diffusion-based super-resolution for two-dimensional spatiotemporal turbulence.
\newblock 433:\penalty0 117478.
\newblock ISSN 0045-7825.
\newblock \doi{10.1016/j.cma.2024.117478}.
\newblock URL \url{https://linkinghub.elsevier.com/retrieve/pii/S0045782524007321}.
\newblock Publisher: Elsevier {BV}.

\bibitem[Fang et~al.(2020)Fang, Sondak, Protopapas, and Succi]{Fang02102020}
Rui Fang, David Sondak, Pavlos Protopapas, and Sauro Succi.
\newblock Neural network models for the anisotropic reynolds stress tensor in turbulent channel flow.
\newblock \emph{Journal of Turbulence}, 21\penalty0 (9-10):\penalty0 525--543, 2020.

\bibitem[Fotiadis et~al.()Fotiadis, Brenowitz, Geffner, Cohen, Pritchard, Vahdat, and Mardani]{fotiadis2025adaptive}
Stathi Fotiadis, Noah~D Brenowitz, Tomas Geffner, Yair Cohen, Michael Pritchard, Arash Vahdat, and Morteza Mardani.
\newblock Adaptive flow matching for resolving small-scale physics.
\newblock In \emph{Forty-second International Conference on Machine Learning}.

\bibitem[Fuoli et~al.(2021)Fuoli, Van~Gool, and Timofte]{fuoli2021fourier}
Dario Fuoli, Luc Van~Gool, and Radu Timofte.
\newblock Fourier space losses for efficient perceptual image super-resolution.
\newblock In \emph{Proceedings of the IEEE/CVF international conference on computer vision}, pages 2360--2369, 2021.

\bibitem[Glawion et~al.(2025)Glawion, Polz, Kunstmann, Fersch, and Chwala]{Glawion2025}
Luca Glawion, Julius Polz, Harald Kunstmann, Benjamin Fersch, and Christian Chwala.
\newblock Global spatio-temporal {{ERA5}} precipitation downscaling to km and sub-hourly scale using generative {{AI}}.
\newblock \emph{npj Climate and Atmospheric Science}, 8\penalty0 (1):\penalty0 219, 2025.

\bibitem[Gutowski et~al.(2020)Gutowski, Ullrich, Hall, Leung, O’Brien, Patricola, Arritt, Bukovsky, Calvin, Feng, Jones, Kooperman, Monier, Pritchard, Pryor, Qian, Rhoades, Roberts, Sakaguchi, Urban, and Zarzycki]{NeedforHighRes}
W.~J. Gutowski, P.~A. Ullrich, A.~Hall, L.~R. Leung, T.~A. O’Brien, C.~M. Patricola, R.~W. Arritt, M.~S. Bukovsky, K.~V. Calvin, Z.~Feng, A.~D. Jones, G.~J. Kooperman, E.~Monier, M.~S. Pritchard, S.~C. Pryor, Y.~Qian, A.~M. Rhoades, A.~F. Roberts, K.~Sakaguchi, N.~Urban, and C.~Zarzycki.
\newblock The ongoing need for high-resolution regional climate models: Process understanding and stakeholder information.
\newblock \emph{Bulletin of the American Meteorological Society}, 101\penalty0 (5):\penalty0 E664 -- E683, 2020.
\newblock \doi{10.1175/BAMS-D-19-0113.1}.
\newblock URL \url{https://journals.ametsoc.org/view/journals/bams/101/5/bams-d-19-0113.1.xml}.

\bibitem[Hakim and Masanam(2024)]{hakim2024}
Gregory~J. Hakim and Sanjit Masanam.
\newblock Dynamical tests of a deep-learning weather prediction model.
\newblock \emph{Artificial Intelligence for the Earth Systems}, May 2024.
\newblock ISSN 2769-7525.
\newblock \doi{10.1175/AIES-D-23-0090.1}.

\bibitem[Harder et~al.(2023)Harder, Hernandez-Garcia, Ramesh, Yang, Sattegeri, Szwarcman, Watson, and Rolnick]{harder2023hard}
Paula Harder, Alex Hernandez-Garcia, Venkatesh Ramesh, Qidong Yang, Prasanna Sattegeri, Daniela Szwarcman, Campbell Watson, and David Rolnick.
\newblock Hard-constrained deep learning for climate downscaling.
\newblock \emph{Journal of Machine Learning Research}, 24\penalty0 (365):\penalty0 1--40, 2023.

\bibitem[Harilal et~al.(2022)Harilal, Hodge, Monteleoni, and Subramanian]{harilal2022enhancedsd}
Nidhin Harilal, Bri-Mathias~S Hodge, Claire Monteleoni, and Aneesh Subramanian.
\newblock Enhanced{{SD}}: Downscaling solar irradiance from climate model projections.
\newblock In \emph{NeurIPS 2022 Workshop on Tackling Climate Change with Machine Learning}, 2022.
\newblock URL \url{https://www.climatechange.ai/papers/neurips2022/67}.

\bibitem[Hersbach(2000)]{hersbach2000decomposition}
Hans Hersbach.
\newblock Decomposition of the continuous ranked probability score for ensemble prediction systems.
\newblock \emph{Weather and Forecasting}, 15\penalty0 (5):\penalty0 559--570, 2000.

\bibitem[Hersbach et~al.(2020)Hersbach, Bell, Berrisford, Hirahara, Hor{\'a}nyi, Mu{\~n}oz-Sabater, Nicolas, Peubey, Radu, Schepers, Simmons, Soci, Abdalla, Abellan, Balsamo, Bechtold, Biavati, Bidlot, Bonavita, Chiara, Dahlgren, Dee, Diamantakis, Dragani, Flemming, Forbes, Fuentes, Geer, Haimberger, Healy, Hogan, H{\'o}lm, Janiskov{\'a}, Keeley, Laloyaux, Lopez, Lupu, Radnoti, Rosnay, Rozum, Vamborg, Villaume, and Th{\'e}paut]{era5}
Hans Hersbach, Bill Bell, Paul Berrisford, Shoji Hirahara, Andr{\'a}s Hor{\'a}nyi, Joaqu{\'i}n Mu{\~n}oz-Sabater, Julien Nicolas, Carole Peubey, Raluca Radu, Dinand Schepers, Adrian Simmons, Cornel Soci, Saleh Abdalla, Xavier Abellan, Gianpaolo Balsamo, Peter Bechtold, Gionata Biavati, Jean Bidlot, Massimo Bonavita, Giovanna Chiara, Per Dahlgren, Dick Dee, Michail Diamantakis, Rossana Dragani, Johannes Flemming, Richard Forbes, Manuel Fuentes, Alan Geer, Leo Haimberger, Sean Healy, Robin~J. Hogan, El{\'i}as H{\'o}lm, Marta Janiskov{\'a}, Sarah Keeley, Patrick Laloyaux, Philippe Lopez, Cristina Lupu, Gabor Radnoti, Patricia Rosnay, Iryna Rozum, Freja Vamborg, Sebastien Villaume, and Jean-No{\"e}l Th{\'e}paut.
\newblock The {{ERA5}} global reanalysis.
\newblock \emph{Quarterly Journal of the Royal Meteorological Society}, 146\penalty0 (730):\penalty0 1999--2049, July 2020.
\newblock ISSN 0035-9009, 1477-870X.
\newblock \doi{10.1002/qj.3803}.

\bibitem[H{\"o}hlein et~al.(2020)H{\"o}hlein, Kern, Hewson, and Westermann]{Hohlein2020WindDownscaling}
Kevin H{\"o}hlein, Michael Kern, Timothy Hewson, and R{\"u}diger Westermann.
\newblock A comparative study of convolutional neural network models for wind field downscaling.
\newblock \emph{Meteorological Applications}, 27\penalty0 (6):\penalty0 e1961, 2020.

\bibitem[Jacobson(2005)]{jacobson2005}
Mark~Z. Jacobson.
\newblock \emph{Fundamentals of Atmospheric Modeling}.
\newblock Cambridge University Press, Cambridge, second edition edition, 2005.
\newblock ISBN 978-1-139-16538-9.
\newblock \doi{10.1017/CBO9781139165389}.

\bibitem[Kajbaf et~al.(2022)Kajbaf, Bensi, and Brubaker]{Kajbaf2022}
Azin~Al Kajbaf, Michelle Bensi, and Kaye~L Brubaker.
\newblock Temporal downscaling of precipitation from climate model projections using machine learning.
\newblock \emph{Stochastic Environmental Research and Risk Assessment}, 36\penalty0 (8):\penalty0 2173--2194, 2022.

\bibitem[Kochkov et~al.()Kochkov, Smith, Alieva, Wang, Brenner, and Hoyer]{kochkov2021}
Dmitrii Kochkov, Jamie~A. Smith, Ayya Alieva, Qing Wang, Michael~P. Brenner, and Stephan Hoyer.
\newblock Machine learning–accelerated computational fluid dynamics.
\newblock 118\penalty0 (21):\penalty0 e2101784118.
\newblock ISSN 0027-8424, 1091-6490.
\newblock \doi{10.1073/pnas.2101784118}.
\newblock URL \url{https://pnas.org/doi/full/10.1073/pnas.2101784118}.

\bibitem[Lam et~al.(2023)Lam, Sanchez-Gonzalez, Willson, Wirnsberger, Fortunato, Alet, Ravuri, Ewalds, Eaton-Rosen, Hu, et~al.]{Sanchez2023}
Remi Lam, Alvaro Sanchez-Gonzalez, Matthew Willson, Peter Wirnsberger, Meire Fortunato, Ferran Alet, Suman Ravuri, Timo Ewalds, Zach Eaton-Rosen, Weihua Hu, et~al.
\newblock Learning skillful medium-range global weather forecasting.
\newblock \emph{Science}, 382\penalty0 (6677):\penalty0 1416--1421, 2023.

\bibitem[Maraun et~al.(2010)Maraun, Wetterhall, Ireson, Chandler, Kendon, Widmann, Brienen, Rust, Sauter, Theme{\ss}l, et~al.]{https://doi.org/10.1029/2009RG000314}
Douglas Maraun, Frederick Wetterhall, Anderson~M Ireson, Richard~E Chandler, Elizabeth~J Kendon, Martin Widmann, S~Brienen, Henning~W Rust, Tobias Sauter, Matthias Theme{\ss}l, et~al.
\newblock Precipitation downscaling under climate change: Recent developments to bridge the gap between dynamical models and the end user.
\newblock \emph{Reviews of geophysics}, 48\penalty0 (3), 2010.

\bibitem[Mardani et~al.(2025)Mardani, Brenowitz, Cohen, Pathak, Chen, Liu, Vahdat, Nabian, Ge, Subramaniam, et~al.]{mardani2025}
Morteza Mardani, Noah Brenowitz, Yair Cohen, Jaideep Pathak, Chieh-Yu Chen, Cheng-Chin Liu, Arash Vahdat, Mohammad~Amin Nabian, Tao Ge, Akshay Subramaniam, et~al.
\newblock Residual corrective diffusion modeling for km-scale atmospheric downscaling.
\newblock \emph{Communications Earth \& Environment}, 6\penalty0 (1):\penalty0 124, 2025.

\bibitem[Merizzi et~al.(2024)Merizzi, Asperti, and Colamonaco]{merizzi2024wind}
Fabio Merizzi, Andrea Asperti, and Stefano Colamonaco.
\newblock Wind speed super-resolution and validation: from {{ERA5}} to {{CERRA}} via diffusion models.
\newblock \emph{Neural Computing and Applications}, 36\penalty0 (34):\penalty0 21899--21921, 2024.

\bibitem[Mills(2005)]{doi:10.1126/science.1112121}
Evan Mills.
\newblock Insurance in a climate of change.
\newblock \emph{Science}, 309\penalty0 (5737):\penalty0 1040--1044, 2005.

\bibitem[Oskarsson et~al.(2024)Oskarsson, Landelius, Deisenroth, and Lindsten]{Oskarsson2024}
Joel Oskarsson, Tomas Landelius, Marc Deisenroth, and Fredrik Lindsten.
\newblock Probabilistic weather forecasting with hierarchical graph neural networks.
\newblock \emph{Advances in Neural Information Processing Systems}, 37:\penalty0 41577--41648, 2024.

\bibitem[{PhysicsNeMo Contributors}(2023)]{physicsnemo2023}
{PhysicsNeMo Contributors}.
\newblock {NVIDIA PhysicsNeMo: An open-source framework for physics-based deep learning in science and engineering}.
\newblock \url{https://github.com/NVIDIA/physicsnemo}, 2023.
\newblock Version released 24 February 2023.

\bibitem[Price et~al.(2025)Price, Sanchez-Gonzalez, Alet, Andersson, El-Kadi, Masters, Ewalds, Stott, Mohamed, Battaglia, et~al.]{Price2025}
Ilan Price, Alvaro Sanchez-Gonzalez, Ferran Alet, Tom~R Andersson, Andrew El-Kadi, Dominic Masters, Timo Ewalds, Jacklynn Stott, Shakir Mohamed, Peter Battaglia, et~al.
\newblock Probabilistic weather forecasting with machine learning.
\newblock \emph{Nature}, 637\penalty0 (8044):\penalty0 84--90, 2025.

\bibitem[Qiu et~al.(2024)Qiu, Khorramfar, Amin, and Howland]{QIU2024100263}
Liying Qiu, Rahman Khorramfar, Saurabh Amin, and Michael~F Howland.
\newblock Decarbonized energy system planning with high-resolution spatial representation of renewables lowers cost.
\newblock \emph{Cell Reports Sustainability}, 1\penalty0 (12), 2024.

\bibitem[Raissi et~al.(2019)Raissi, Perdikaris, and Karniadakis]{raissi2019}
M.~Raissi, P.~Perdikaris, and G.E. Karniadakis.
\newblock Physics-informed neural networks: {{A}} deep learning framework for solving forward and inverse problems involving nonlinear partial differential equations.
\newblock \emph{Journal of Computational Physics}, 378:\penalty0 686--707, February 2019.
\newblock ISSN 00219991.
\newblock \doi{10.1016/j.jcp.2018.10.045}.

\bibitem[Salath{\'e}~Jr et~al.(2010)Salath{\'e}~Jr, Leung, Qian, and Zhang]{salathe2010regional}
Eric~P Salath{\'e}~Jr, L~Ruby Leung, Yun Qian, and Yongxin Zhang.
\newblock Regional climate model projections for the state of washington.
\newblock \emph{Climatic Change}, 102\penalty0 (1):\penalty0 51--75, 2010.

\bibitem[Schimanke et~al.(2021)Schimanke, Ridal, Le~Moigne, Berggren, Und{\'e}n, Randriamampianina, Andrea, Bazile, Bertelsen, Brousseau, Dahlgren, Edvinsson, El~Said, Glinton, Hopsch, Isaksson, Mladek, Olsson, Verrelle, and Wang]{cerra}
S.~Schimanke, M.~Ridal, P.~Le~Moigne, L.~Berggren, P.~Und{\'e}n, R.~Randriamampianina, U.~Andrea, E.~Bazile, A.~Bertelsen, P.~Brousseau, P.~Dahlgren, L.~Edvinsson, A.~El~Said, M.~Glinton, S.~Hopsch, L.~Isaksson, R.~Mladek, E.~Olsson, A.~Verrelle, and Z.Q. Wang.
\newblock {{CERRA}} sub-daily regional reanalysis data for {{Europe}} on single levels from 1984 to present, 2021.

\bibitem[Sekiyama et~al.(2023)Sekiyama, Hayashi, Kaneko, and Fukui]{sekiyama2023}
Tsuyoshi~Thomas Sekiyama, Syugo Hayashi, Ryo Kaneko, and Ken-ichi Fukui.
\newblock Surrogate downscaling of mesoscale wind fields using ensemble superresolution convolutional neural networks.
\newblock \emph{Artificial Intelligence for the Earth Systems}, 2\penalty0 (3), July 2023.
\newblock ISSN 2769-7525.
\newblock \doi{10.1175/aies-d-23-0007.1}.

\bibitem[Selz and Craig(2023)]{selz2023}
T.~Selz and G.~C. Craig.
\newblock Can artificial intelligence-based weather prediction models simulate the butterfly effect?
\newblock \emph{Geophysical Research Letters}, 50\penalty0 (20):\penalty0 e2023GL105747, 2023.
\newblock ISSN 1944-8007.
\newblock \doi{10.1029/2023GL105747}.

\bibitem[Skamarock et~al.(2021)Skamarock, Klemp, Dudhia, Gill, Liu, Berner, Wang, Powers, Duda, Barker, and Huang]{skamarock2021}
William~C. Skamarock, Joseph~B. Klemp, Jimy Dudhia, David~O. Gill, Zhiquan Liu, Judith Berner, Wei Wang, Jordan~G. Powers, Michael~G. Duda, Dale~M. Barker, and Xiang-Yu Huang.
\newblock A description of the {{Advanced Research WRF Model Version}} 4.3.
\newblock Technical report, UCAR/NCAR, July 2021.

\bibitem[Tomasi et~al.(2025)Tomasi, Franch, and Cristoforetti]{gmd-18-2051-2025}
Elena Tomasi, Gabriele Franch, and Marco Cristoforetti.
\newblock Can ai be enabled to perform dynamical downscaling? a latent diffusion model to mimic kilometer-scale cosmo5. 0\_clm9 simulations.
\newblock \emph{Geoscientific Model Development}, 18\penalty0 (6):\penalty0 2051--2078, 2025.

\bibitem[Wilby et~al.(1998)Wilby, Wigley, Conway, Jones, Hewitson, Main, and Wilks]{Wilby1998}
Robert~L Wilby, TML Wigley, D~Conway, PD~Jones, BC~Hewitson, J~Main, and DS~Wilks.
\newblock Statistical downscaling of general circulation model output: A comparison of methods.
\newblock \emph{Water resources research}, 34\penalty0 (11):\penalty0 2995--3008, 1998.

\bibitem[Yan et~al.(2024)Yan, Foo, Trinh, Yeung, Wong, and Wong]{yan2024fourier}
Chiu-Wai Yan, Shi~Quan Foo, Van~Hoan Trinh, Dit-Yan Yeung, Ka-Hing Wong, and Wai-Kin Wong.
\newblock Fourier amplitude and correlation loss: Beyond using l2 loss for skillful precipitation nowcasting.
\newblock \emph{Advances in Neural Information Processing Systems}, 37:\penalty0 100007--100041, 2024.

\end{thebibliography}
%------------------------------------------------------------------------------------------------------------------------------
%%%%% Appendix %%%%%%%%%%%
%------------------------------------------------------------------------------------------------------------------------------
\newpage
\appendix
\section*{Appendix} % Standalone heading
\addcontentsline{toc}{section}{Appendix} % Optional: add to TOC
\section{Fourier Differentiation Amplifies High Frequencies}
\label{app:fourier_derivative}

\paragraph{Fourier transform and wavenumbers.}
Let \(\mathbf{q} \in \mathbb{R}^{H \times W}\) be a real-valued scalar field defined on a uniform \(H \times W\) grid with isotropic grid spacing \(\Delta x\). We denote its 2D discrete Fourier transform by \(\mathcal{F}(\mathbf{q})(k_h, k_w)\), where \(k_h\) and \(k_w\) are discrete mode indices. The corresponding physical wavenumbers are
\[
\kappa_x = \frac{2\pi k_w}{W\Delta x}, \qquad\text{and}\qquad
\kappa_y = \frac{2\pi k_h}{H\Delta x}.
\]
We define the horizontal wavenumber vector \(\boldsymbol{\kappa} = (\kappa_x, \kappa_y)\) and its magnitude \(\kappa = \|\boldsymbol{\kappa}\| = \sqrt{\kappa_x^2 + \kappa_y^2}\). High values of \(\kappa\) correspond to high-frequency, small-scale features.
%\paragraph{Power spectral density (PSD).}
Then, the discrete PSD of \(\mathbf{q}\) is given by
\[
\mathrm{PSD}(\mathbf{q})(k_h,k_w) = \frac{\bigl|\mathcal{F}(\mathbf{q})(k_h,k_w)\bigr|^2}{H\,W\,\Delta x}.
\]
This quantity measures the contribution of each wavenumber mode to the field's total variance.

%\paragraph{Fourier differentiation rule.}
Now we examine how spatial derivatives affect its spectral content.The discrete Fourier transform of a spatial derivative is obtained by multiplying by \(i\) times the corresponding wavenumber:
\[
\mathcal{F}\!\left(\frac{\partial \mathbf{q}}{\partial x}\right) = i\,\kappa_x\,\mathcal{F}(\mathbf{q}),
\qquad\text{and}\qquad
\mathcal{F}\!\left(\frac{\partial \mathbf{q}}{\partial y}\right) = i\,\kappa_y\,\mathcal{F}(\mathbf{q}).
\]
Therefore, the PSDs of the derivatives are
\[
\mathrm{PSD}\!\left(\frac{\partial \mathbf{q}}{\partial x}\right) = \kappa_x^2\,\mathrm{PSD}(\mathbf{q}),
\qquad\text{and}\qquad
\mathrm{PSD}\!\left(\frac{\partial \mathbf{q}}{\partial y}\right) = \kappa_y^2\,\mathrm{PSD}(\mathbf{q}).
\]
Consequently, for the horizontal gradient \(\nabla \mathbf{q} = (\partial_x \mathbf{q},\, \partial_y \mathbf{q})\), we have
\[
\bigl\|\mathcal{F}(\nabla \mathbf{q})\bigr\|^2 = \left\Vert \begin{bmatrix}
    \mathcal{F}\!\left(\frac{\partial \mathbf{q}}{\partial x}\right)  & \mathcal{F}\!\left(\frac{\partial \mathbf{q}}{\partial y}\right)
\end{bmatrix} \right\Vert^2 = \kappa^2\,\bigl|\mathcal{F}(\mathbf{q})\bigr|^2,
\]
so that the corresponding PSD is
\[
%\bigl\|\mathcal{F}(\nabla \mathbf{q})\bigr\|^2 = \kappa^2\,\bigl|\mathcal{F}(\mathbf{q})\bigr|^2 
%\quad\Longrightarrow\quad
\mathrm{PSD}(\nabla \mathbf{q})  = \kappa^2\,\mathrm{PSD}(\mathbf{q}).
\]
Since \(\kappa\) increases with frequency, the \(\kappa^2\) factor explicitly shows that high-frequency modes are multiplied by larger factors, meaning they are \emph{amplified}.

\paragraph{Example: horizontal divergence.}
Let \(\boldsymbol{\delta}_h\) be the horizontal divergence defined as
\[
\boldsymbol{\delta}_h = \nabla_h \cdot \mathbf{v}_h = \frac{\partial \mathbf{u}}{\partial x} + \frac{\partial \mathbf{v}}{\partial y}.
\]
In Fourier space,
\[
\mathcal{F}(\boldsymbol{\delta}_h) = i\,\kappa_x\,\mathcal{F}(\mathbf{u}) + i\,\kappa_y\,\mathcal{F}(\mathbf{v}).
\]
Its power spectrum is
\[
\bigl|\mathcal{F}(\boldsymbol{\delta}_h)\bigr|^2
= \kappa_x^2\,\bigl|\mathcal{F}(\mathbf{u})\bigr|^2
+ \kappa_y^2\,\bigl|\mathcal{F}(\mathbf{v})\bigr|^2
+ 2\,\kappa_x\kappa_y\,\Re\!\bigl(\mathcal{F}(\mathbf{u})\,\mathcal{F}(\mathbf{v})^{*}\bigr).
\]
The leading \(\kappa_x^2\) and \(\kappa_y^2\) factors again make the amplification of high-\(\kappa\) modes explicit, demonstrating why high-frequency modes are strongly amplified.

Differentiation in Fourier space multiplies each mode by its wavenumber magnitude \(\kappa\), so the power in derivative-based diagnostics is weighted by \(\kappa^2\).  
For the horizontal wind components \(\mathbf{u}\) and \(\mathbf{v}\), any loss of high-frequency variance in \(\mathrm{PSD}(\mathbf{u})\) or \(\mathrm{PSD}(\mathbf{v})\) is magnified in the spectra of \(\boldsymbol{\delta}_h\) and \(\boldsymbol{\zeta}_h\), since their definitions involve first-order derivatives in both spatial directions.  
Consequently, models that over-smooth \(\mathbf{u}\) and \(\mathbf{v}\) will not only misrepresent their own high-\(\kappa\) PSDs, but will show even larger discrepancies in the PSDs of \(\boldsymbol{\delta}_h\) and \(\boldsymbol{\zeta}_h\).  
By explicitly constraining the model to match the full scale-dependent variance of \(\mathbf{u}\) and \(\mathbf{v}\) through a PSD-based loss, we reduce this high-frequency deficit and thereby improve the physical consistency of all derivative-based diagnostics.

\section{Dataset and training details}
\label{section:Dataset details}

ERA5 \citep{era5} and CERRA \cite{cerra} are both multi-decade reanalysis products produced by the European Centre for Medium-Range Weather Forecasts (ECMWF) using state-of-the-art numerical weather models with data assimilation. 
CERRA provides high-resolution fields over Europe at 5.5~km grid spacing and a temporal resolution of three hours, whereas ERA5 covers the globe at 25~km resolution and hourly intervals. 
This corresponds to a spatial super-resolution factor of approximately \(4.54\times\) in each horizontal direction. 
Since the downscaling task requires temporally aligned input–target pairs, we sub-sample ERA5 to every third hour to match the temporal availability of CERRA.

\paragraph{Training, validation, and test splits.}  
For training, we select a Central European domain bounded by \([60^\circ\mathrm{N}, -2^\circ\mathrm{E}, 40^\circ\mathrm{N}, 18^\circ\mathrm{E}]\)
The training period spans from 1~January~2014 to 31~December~2020.  
We reserve the year~2021 for validation and in-distribution testing: January, March, May, July, September, and November are used for testing, while the remaining months serve as the validation set.

\paragraph{Geographic generalization experiments.}  
To evaluate cross-regional generalization, we use the same 2021 monthly split but replace the Central European domain with two non-overlapping target regions:
\begin{itemize}
    \item \textbf{Iberia, Morocco:} \([45^\circ\mathrm{N}, -15^\circ\mathrm{E}, 25^\circ\mathrm{N}, 5^\circ\mathrm{E}]\)
    \item \textbf{Northern Scandinavia:} \([70^\circ\mathrm{N}, 24^\circ\mathrm{E}, 50^\circ\mathrm{N}, 44^\circ\mathrm{E}]\)
\end{itemize}
These domains differ substantially in climate and topography from the training region, providing a robust test of spatial transferability.

\paragraph{Training details.}  
We train \textit{CorrDiff} and its regression component, \textit{Regression-CorrDiff}, by adapting the original \emph{PhysicsNeMo} code repository provided by the authors \citep{physicsnemo2023}. 
To train the CRPS-based U-Net ensemble, we follow the methodology of \citet{alet2025skillfuljointprobabilisticweather}, but still adapt the U-Net architecture from the \emph{PhysicsNeMo} repository to construct the ensemble. 
Unlike \citet{alet2025skillfuljointprobabilisticweather}, who train four models (M1, M2, M3, M4) each with its own ensemble, we restrict our experiments to a single model configuration (M1).

\newpage

%------------------------------------------------------------------------------------------------------------------------------
%%%%% RESULTS %%%%%%%%%%%
%------------------------------------------------------------------------------------------------------------------------------
\section{Results Central Europe}
\label{app:results-CE}
%%%%%%%%%%% PLOTTED results - U wind - Central Europe
\begin{figure}[htbp!]
  \centering
  % Row 1: two centered images
  \begin{subfigure}[t]{0.23\textwidth}
    \centering
    \includegraphics[width=\linewidth]{Images/predsCE/u/ERA5.png}
    \subcaption{Input: ERA5}
    
  \end{subfigure}
  \hspace{0.04\textwidth} % small gap to center two images
  \begin{subfigure}[t]{0.23\textwidth}
    \centering
    \includegraphics[width=\linewidth]{Images/predsCE/u/CERRA.png}
    \subcaption{Target: CERRA}
    
  \end{subfigure}

  \vspace{0.8em}

  % Row 2: three equally spaced images
  \begin{subfigure}[t]{0.23\textwidth}
    \centering
    \includegraphics[width=\linewidth]{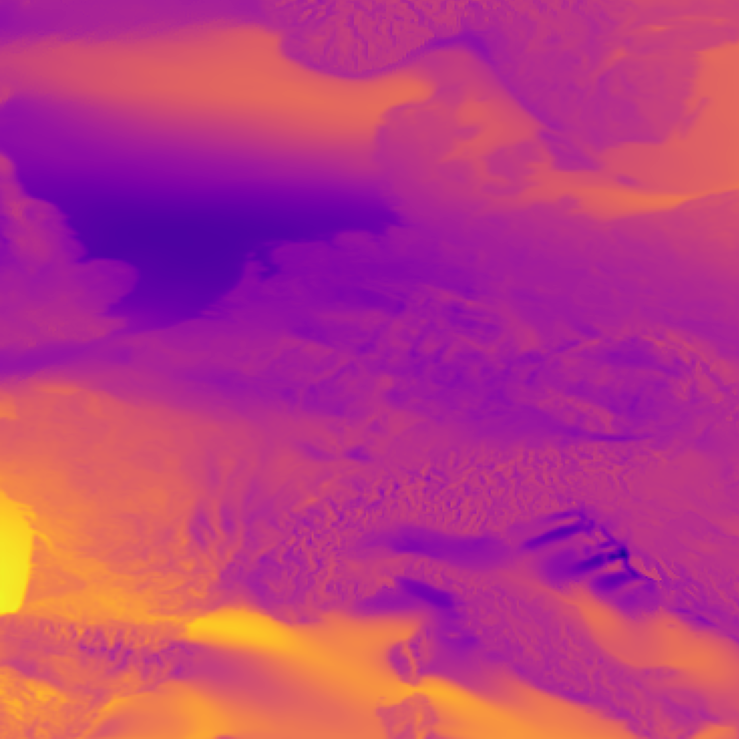}
    \subcaption{CorrDiff}
    
  \end{subfigure}
  \hfill
  \begin{subfigure}[t]{0.23\textwidth}
    \centering
    \includegraphics[width=\linewidth]{Images/predsCE/u/CRPS-PSD.png}
    \subcaption{CRPS UNets-PSD}
    
  \end{subfigure}
  \hfill
  \begin{subfigure}[t]{0.23\textwidth}
    \centering
    \includegraphics[width=\linewidth]{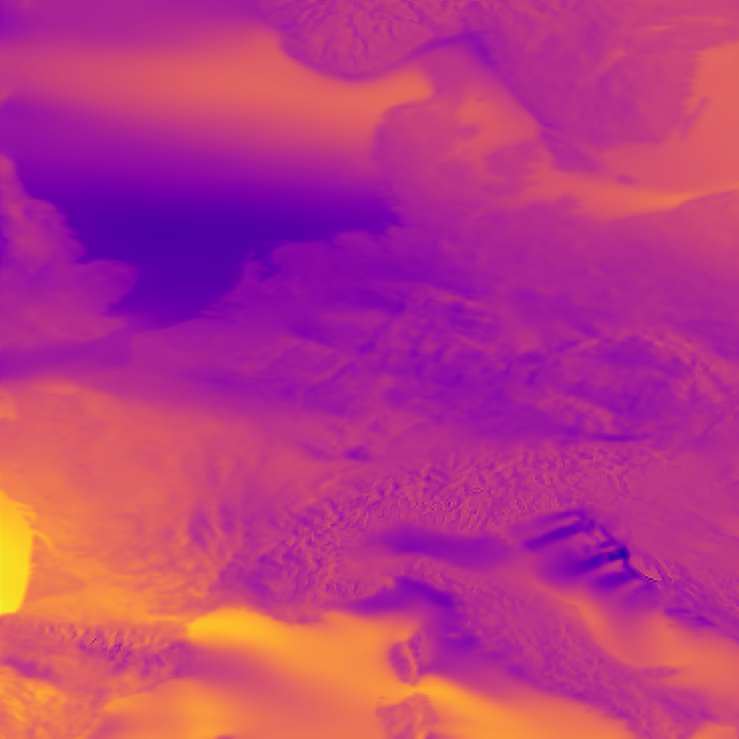}
    \subcaption{Reg-CorrDiff-PSD}
    
  \end{subfigure}

  \vspace{0.8em}

  % Row 3: full-width colormap
  \begin{subfigure}[t]{0.41\textwidth}
    \centering
    \includegraphics[width=\linewidth]{Images/predsCE/u/colorbar.png}
    
  \end{subfigure}
\vspace{-0.9em}
  \caption{Input, target, and predictions of $\mathbf{u}$ in Central Europe (in-distribution) using \textit{CorrDiff}, \textit{Reg-CorrDiff-PSD}, and \textit{CRPS UNets-PSD}. \textit{CorrDiff} yields a visually sharper prediction, while the others appear slightly blurrier.}
  \label{fig:preds-u-appendix-CE}
\end{figure}

%%%%%%%%%%% PLOTTED results - U wind - Central Europe
\begin{figure}[htbp!]
  \centering
  % Row 1: two centered images
  \begin{subfigure}[t]{0.23\textwidth}
    \centering
    \includegraphics[width=\linewidth]{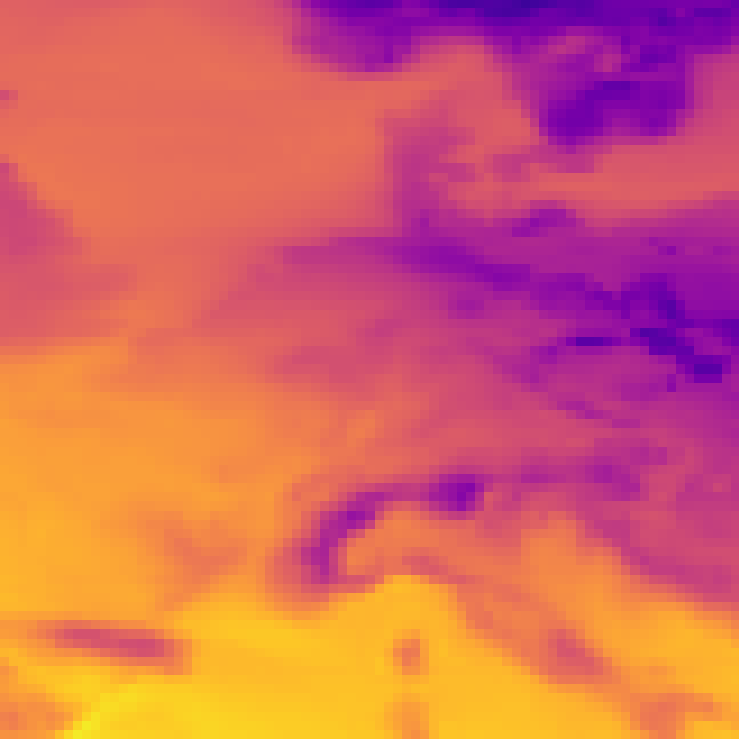}
    \subcaption{Input: ERA5}
  \end{subfigure}
  \hspace{0.04\textwidth} % small gap to center two images
  \begin{subfigure}[t]{0.23\textwidth}
    \centering
    \includegraphics[width=\linewidth]{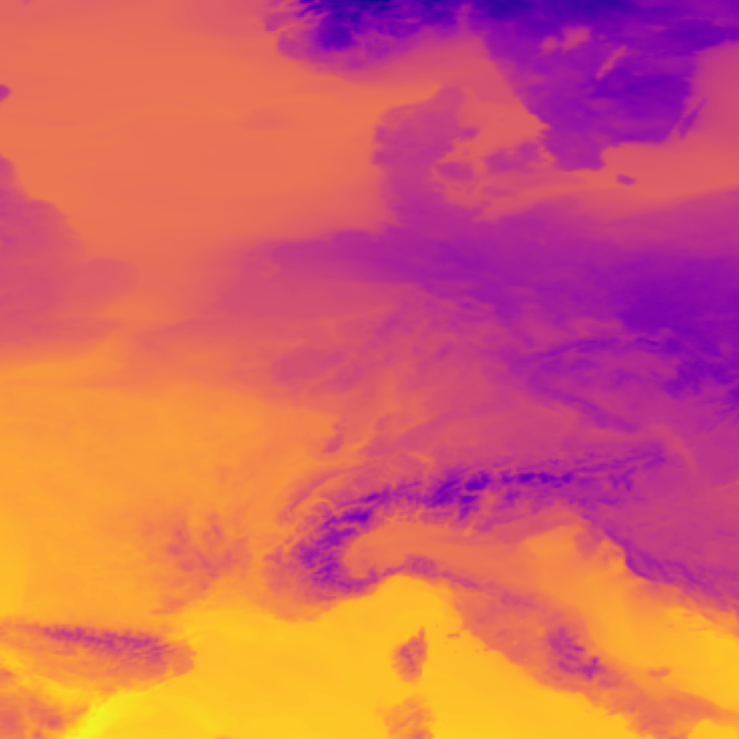}
    \subcaption{Target: CERRA}
    
  \end{subfigure}

  \vspace{0.8em}

  % Row 2: three equally spaced images
  \begin{subfigure}[t]{0.23\textwidth}
    \centering
    \includegraphics[width=\linewidth]{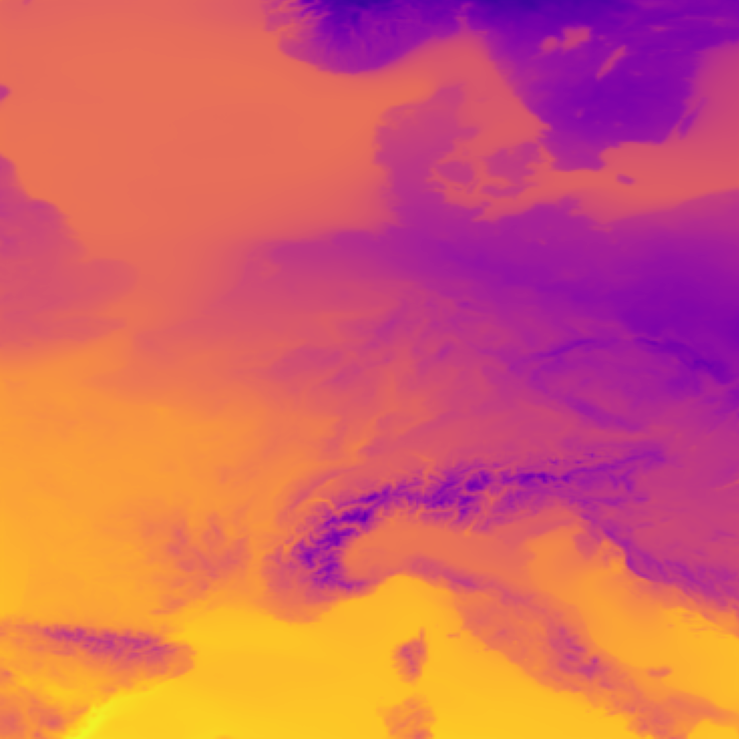}
    \subcaption{CorrDiff}
    
  \end{subfigure}
  \hfill
  \begin{subfigure}[t]{0.23\textwidth}
    \centering
    \includegraphics[width=\linewidth]{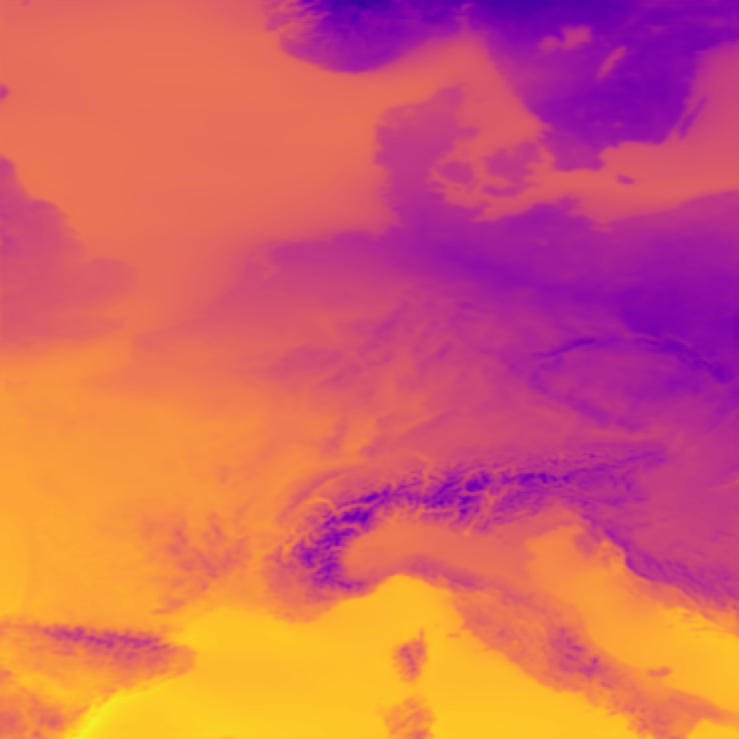}
    \subcaption{CRPS UNets-PSD}
    
  \end{subfigure}
  \hfill
  \begin{subfigure}[t]{0.24\textwidth}
    \centering
    \includegraphics[width=\linewidth]{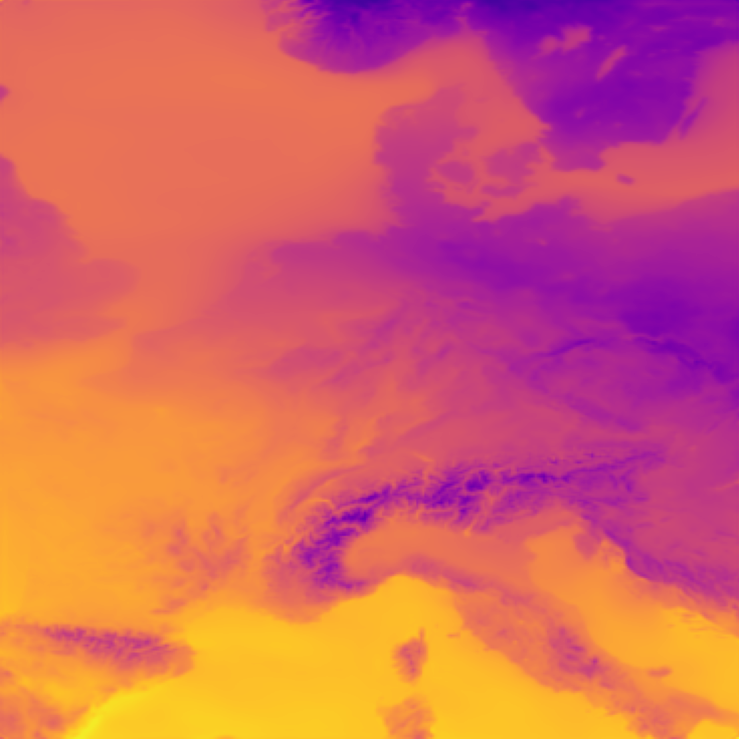}
    \subcaption{Reg-CorrDiff-PSD}
    
  \end{subfigure}

  \vspace{0.8em}

  % Row 3: full-width colormap
  \begin{subfigure}[t]{0.41\textwidth}
    \centering
    \includegraphics[width=\linewidth]{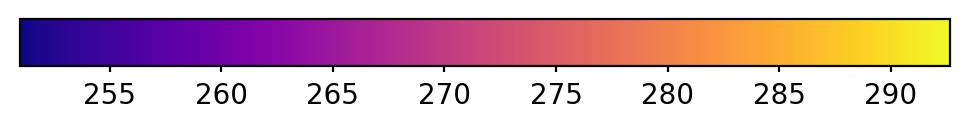}
    
  \end{subfigure}
\vspace{-0.9em}
  \caption{Input, target, and predictions of $\mathbf{t2m}$ in Central Europe (in-distribution) using \textit{CorrDiff}, \textit{Reg-CorrDiff-PSD}, and \textit{CRPS UNets-PSD}. $\mathbf{t2m}$ is easier to predict, hence differences between outputs are hard to spot visually.}
  \label{fig:preds-t-appendix-CE}
\end{figure}

\newpage
%%%%%%%%%%%%%%%%%%%%%%%%% IBERIA %%%%%%%%%%
\vspace{-2.3pt}
\section{Results Iberia, Morocco}
\label{app:results-Iberia}
\vspace{-2.3pt}

%%%%%%%%%%% PLOTTED results - U wind - Iberia
\begin{figure}[htbp!]
  \centering
  % Row 1: two centered images
  \begin{subfigure}[htbp!]{0.23\textwidth}
    \centering
    \includegraphics[width=\linewidth]{Images/predsIberia/u/ERA5.png}
    \subcaption{Input: ERA5}
    
  \end{subfigure}
  \hspace{0.04\textwidth} % small gap to center two images
  \begin{subfigure}[htbp!]{0.23\textwidth}
    \centering
    \includegraphics[width=\linewidth]{Images/predsIberia/u/CERRA.png}
    \subcaption{Target: CERRA}
    
  \end{subfigure}

  \vspace{0.3em}

  % Row 2: three equally spaced images
  \begin{subfigure}[t]{0.23\textwidth}
    \centering
    \includegraphics[width=\linewidth]{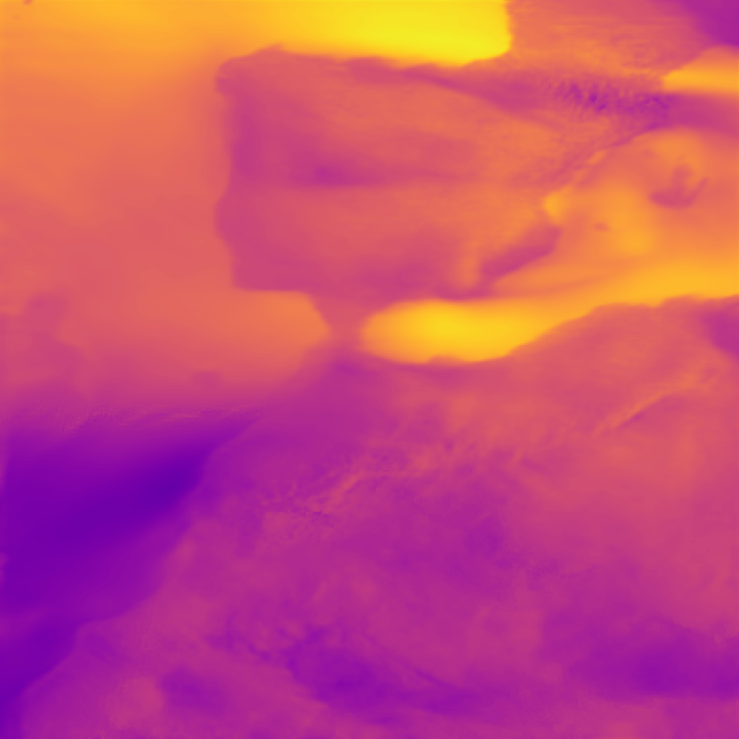}
    \subcaption{CorrDiff}
    
  \end{subfigure}
  \hfill
  \begin{subfigure}[t]{0.23\textwidth}
    \centering
    \includegraphics[width=\linewidth]{Images/predsIberia/u/CRPS-PSD.png}
    \subcaption{CRPS UNets-PSD}
    
  \end{subfigure}
  \hfill
  \begin{subfigure}[t]{0.23\textwidth}
    \centering
    \includegraphics[width=\linewidth]{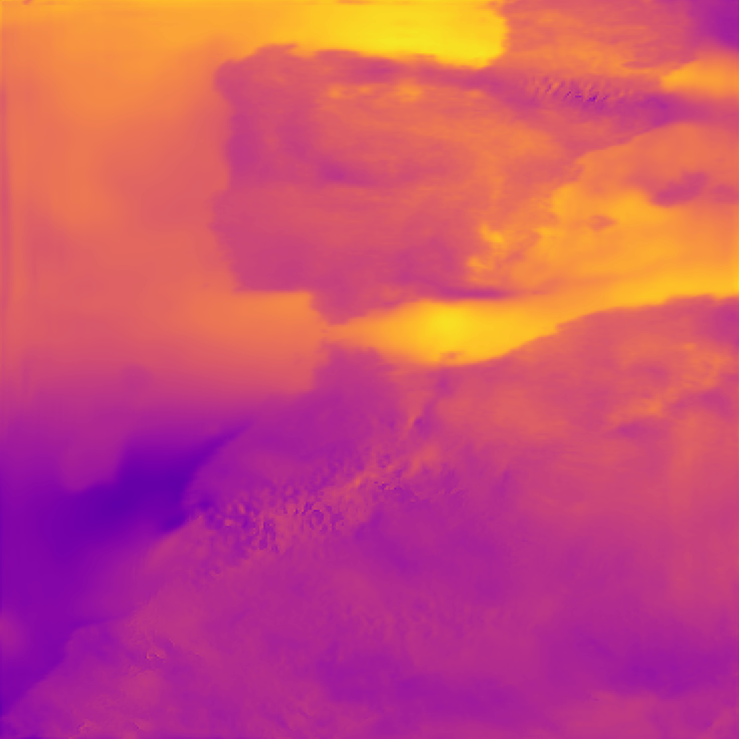}
    \subcaption{Reg-CorrDiff-PSD}
    
  \end{subfigure}

  \vspace{0.3em}

  % Row 3: full-width colormap
  \begin{subfigure}[t]{0.41\textwidth}
    \centering
    \includegraphics[width=\linewidth]{Images/predsIberia/u/colorbar.png}
    
  \end{subfigure}

   \vspace{0.3em}

  \caption{Input, target, and predictions of $\mathbf{u}$ in Iberia, Morocco (OOD) using \textit{CorrDiff}, \textit{Reg-CorrDiff-PSD}, and \textit{CRPS UNets-PSD}. \textit{CRPS UNets-PSD} yields a visually sharper prediction, but still significantly worse compared to the in-distribution scenario.}
  \label{fig:qualitative-grid}
\end{figure}

%%%%%%%%%%% PLOTTED results - T2m wind - Iberia
\begin{figure}[htbp!]
  \centering
  % Row 1: two centered images
  \begin{subfigure}[htbp!]{0.23\textwidth}
    \centering
    \includegraphics[width=\linewidth]{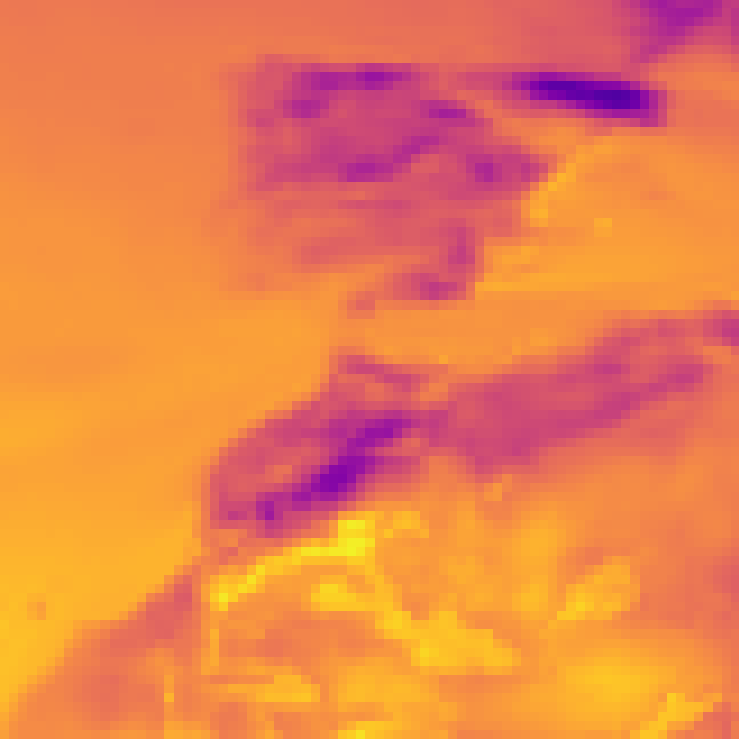}
    \subcaption{Input: ERA5}
    
  \end{subfigure}
  \hspace{0.04\textwidth} % small gap to center two images
  \begin{subfigure}[htbp!]{0.23\textwidth}
    \centering
    \includegraphics[width=\linewidth]{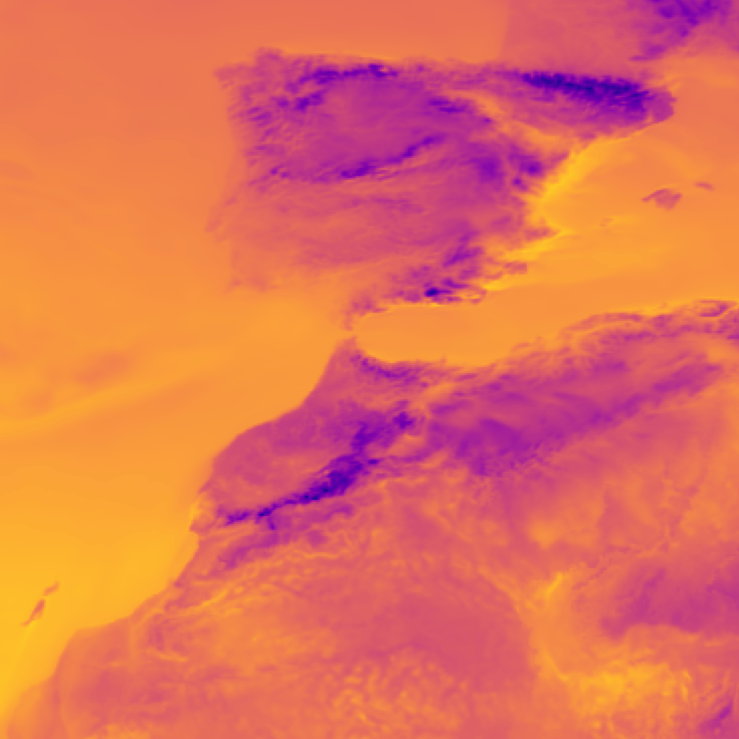}
    \subcaption{Target: CERRA}
    
  \end{subfigure}

  \vspace{0.3em}

  % Row 2: three equally spaced images
  \begin{subfigure}[t]{0.23\textwidth}
    \centering
    \includegraphics[width=\linewidth]{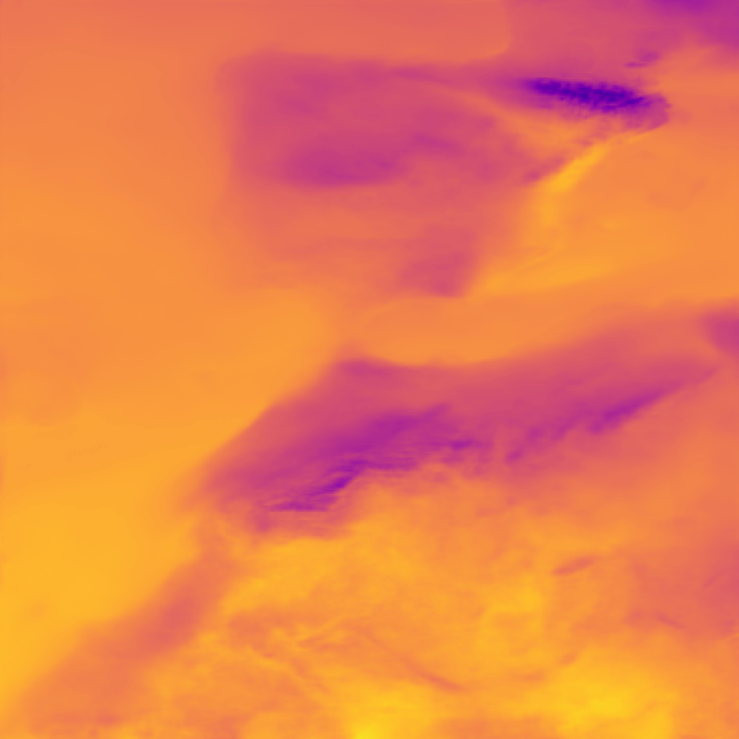}
    \subcaption{CorrDiff}
    
  \end{subfigure}
  \hfill
  \begin{subfigure}[t]{0.23\textwidth}
    \centering
    \includegraphics[width=\linewidth]{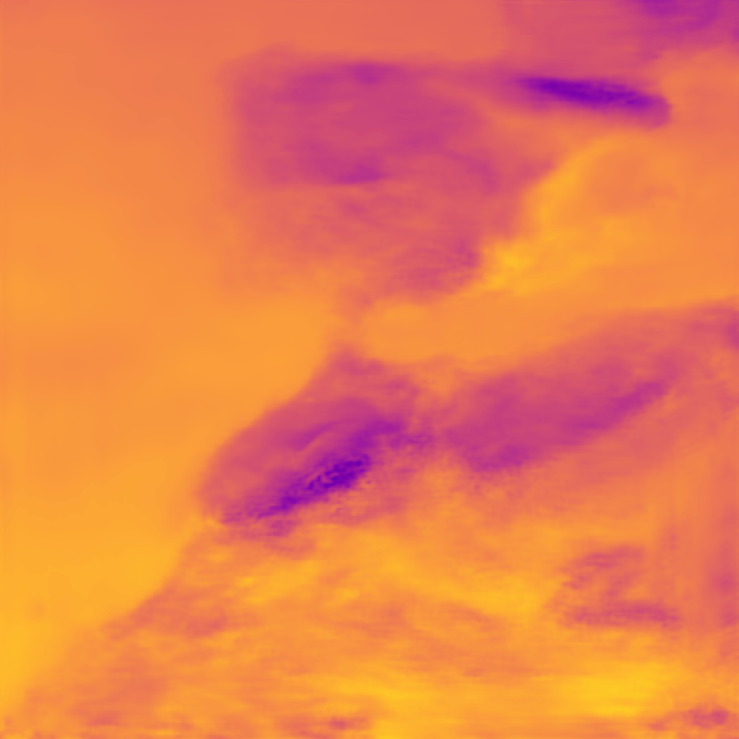}
    \subcaption{CRPS UNets-PSD}
    
  \end{subfigure}
  \hfill
  \begin{subfigure}[t]{0.23\textwidth}
    \centering
    \includegraphics[width=\linewidth]{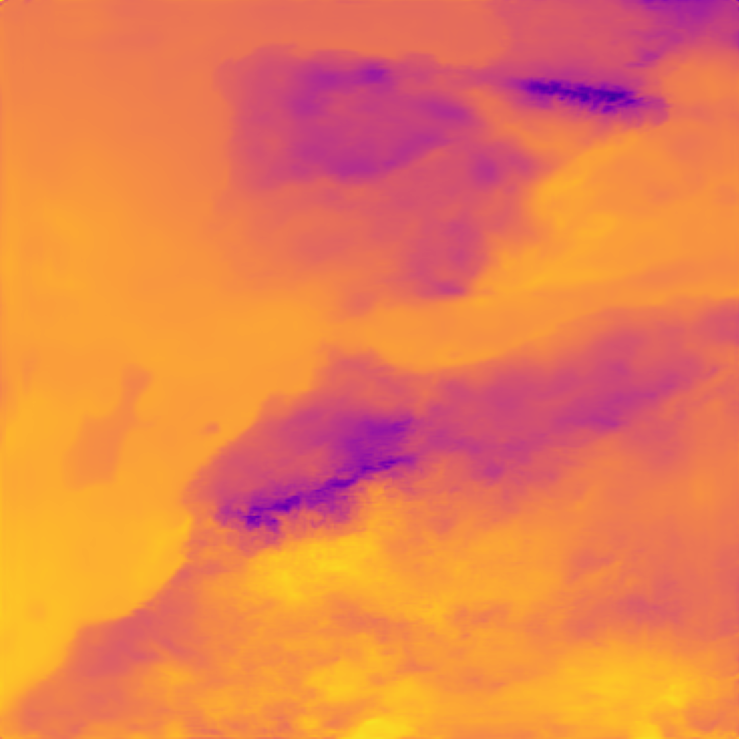}
    \subcaption{Reg-CorrDiff-PSD}
    
  \end{subfigure}

  \vspace{0.3em}

  % Row 3: full-width colormap
  \begin{subfigure}[t]{0.41\textwidth}
    \centering
    \includegraphics[width=\linewidth]{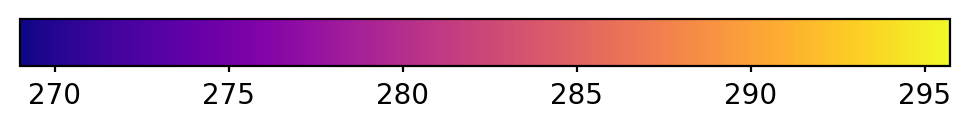}
    
  \end{subfigure}

   \vspace{0.3em}

  \caption{Input, target, and predictions of $\mathbf{t2m}$ in Iberia, Morocco (OOD) using \textit{CorrDiff}, \textit{Reg-CorrDiff-PSD}, and \textit{CRPS UNets-PSD}. \textit{CRPS UNets-PSD} yields a visually sharper prediction, but still significantly worse compared to the in-distribution scenario.}
  \label{fig:qualitative-grid}
\end{figure}

%%%%%%%%%%% PSD IBERIA, DOWNSCALED VARIABLES
\begin{figure}[htbp!]
  \centering
  \includegraphics[width=0.6\linewidth]{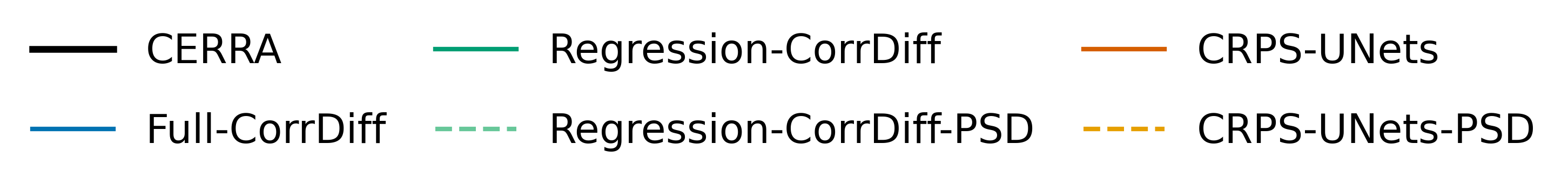}
\end{figure}
\vspace{-8pt}
%%%%%%%%%%% PSD IBERIA – ALL VARIABLES IN ONE FIGURE
\begin{figure}[htbp!]
  \centering
  %------------------ 1st ROW ------------------
  \begin{subfigure}[t]{0.31\textwidth}
    \centering
    \includegraphics[width=\linewidth]{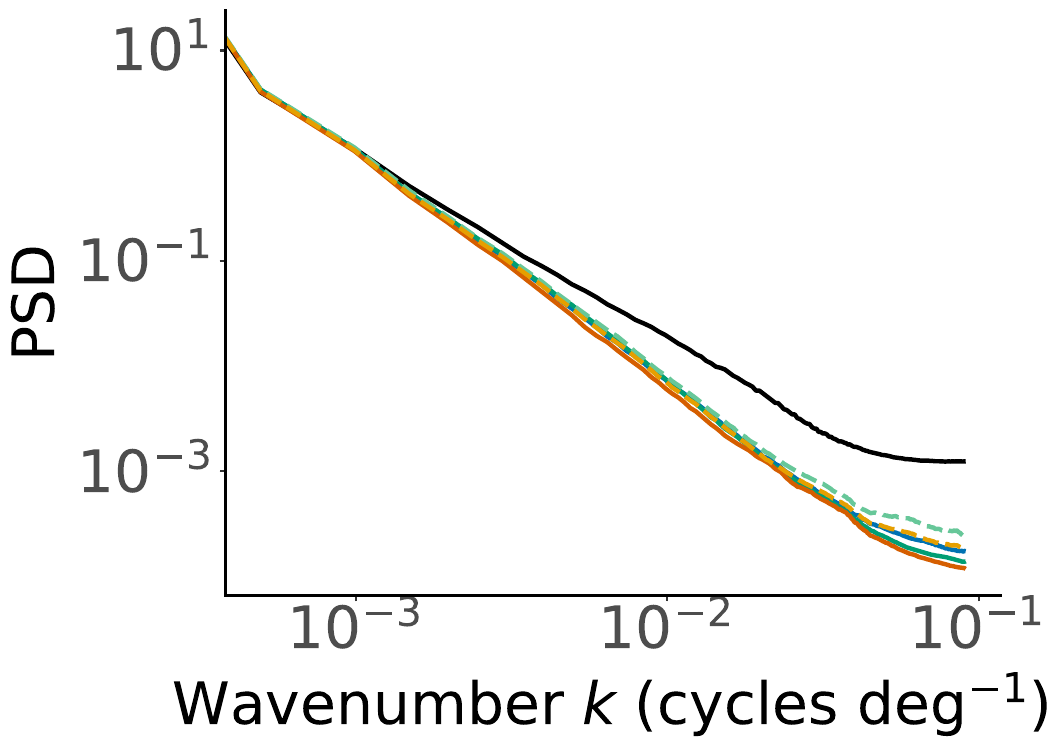}
    \caption{PSD of $\mathbf{u}$}
  \end{subfigure}\hfill
  \begin{subfigure}[t]{0.31\textwidth}
    \centering
    \includegraphics[width=\linewidth]{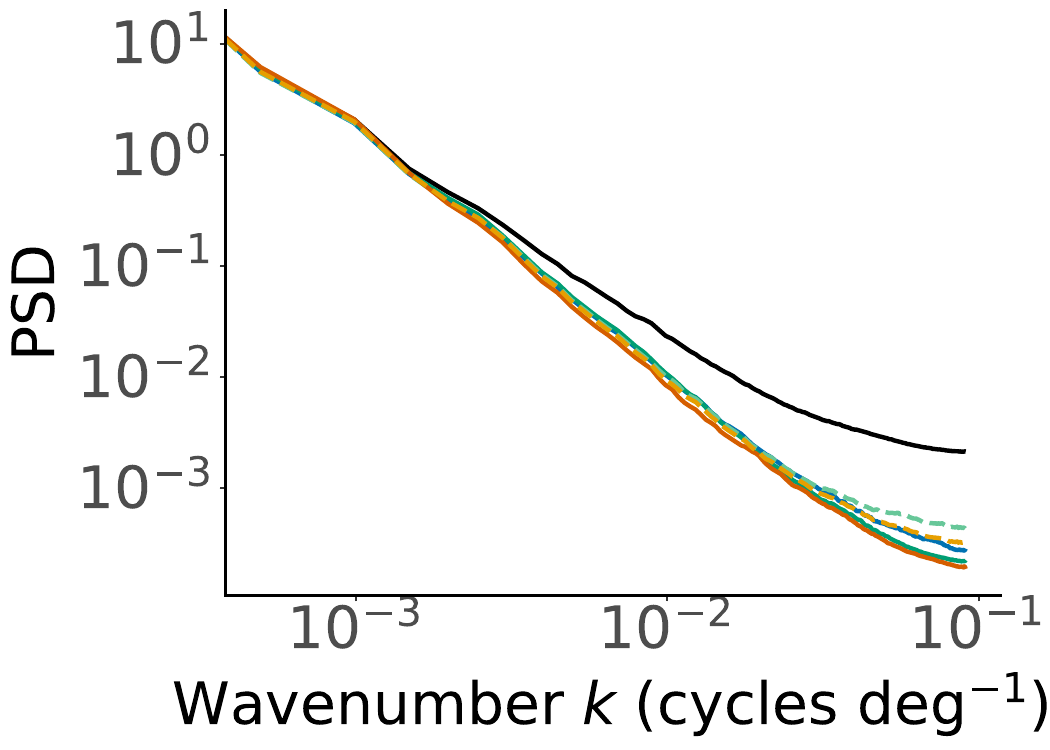}
    \caption{PSD of $\mathbf{v}$}
  \end{subfigure}\hfill
  \begin{subfigure}[t]{0.31\textwidth}
    \centering
    \includegraphics[width=\linewidth]{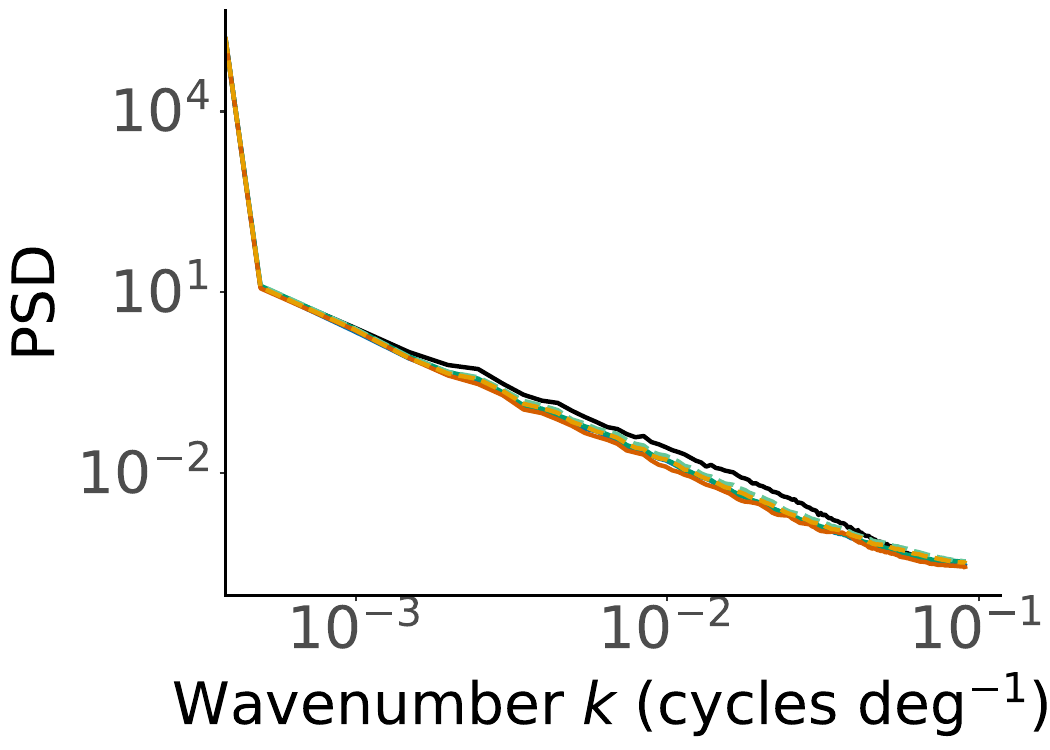}
    \caption{PSD of temperature $\mathbf{t2m}$}
  \end{subfigure}

  \vspace{0.3em} % vertical gap between rows
  
  %------------------ 2nd ROW ------------------
  \begin{subfigure}[t]{0.31\textwidth}
    \centering
    \includegraphics[width=\linewidth]{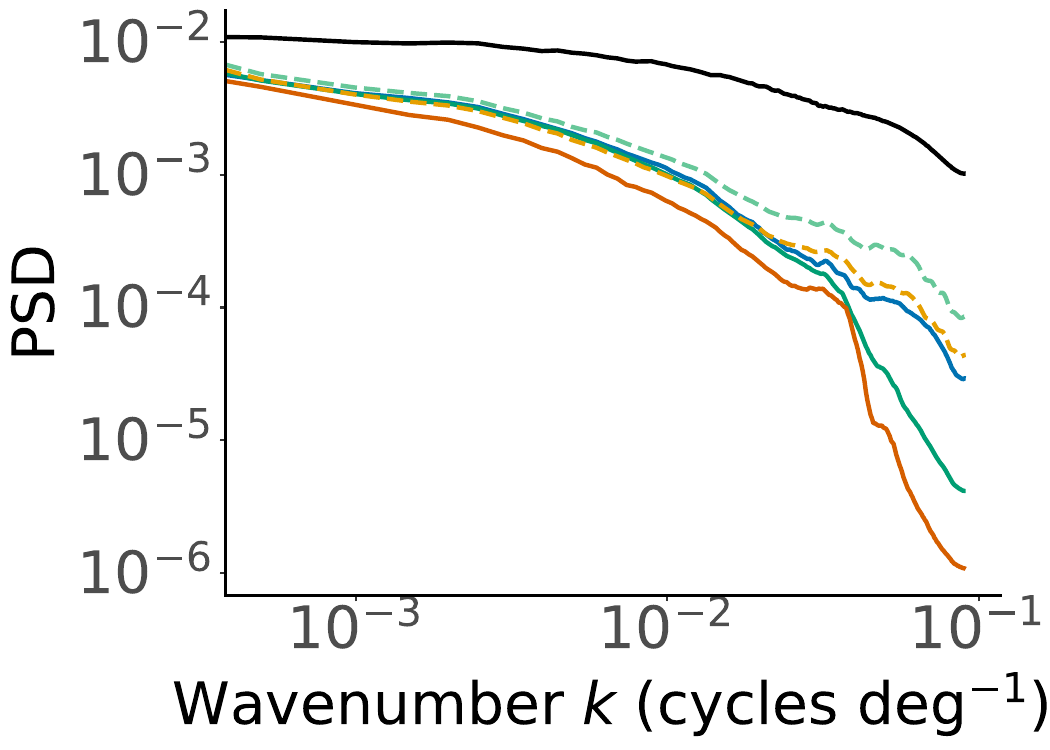}
    \caption{PSD of $\boldsymbol{\delta}_h$}
  \end{subfigure}\hfill
  \begin{subfigure}[t]{0.31\textwidth}
    \centering
    \includegraphics[width=\linewidth]{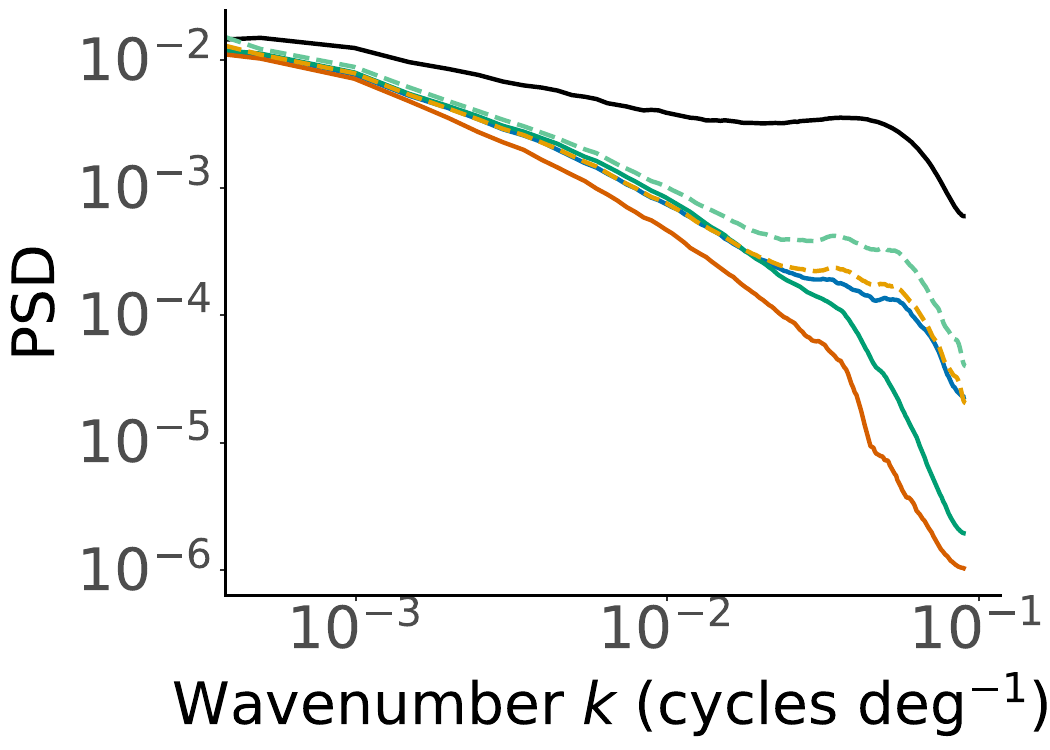}
    \caption{PSD of $\boldsymbol{\zeta}_h$}
  \end{subfigure}\hfill
  \begin{subfigure}[t]{0.31\textwidth}
    \centering
    \includegraphics[width=\linewidth]{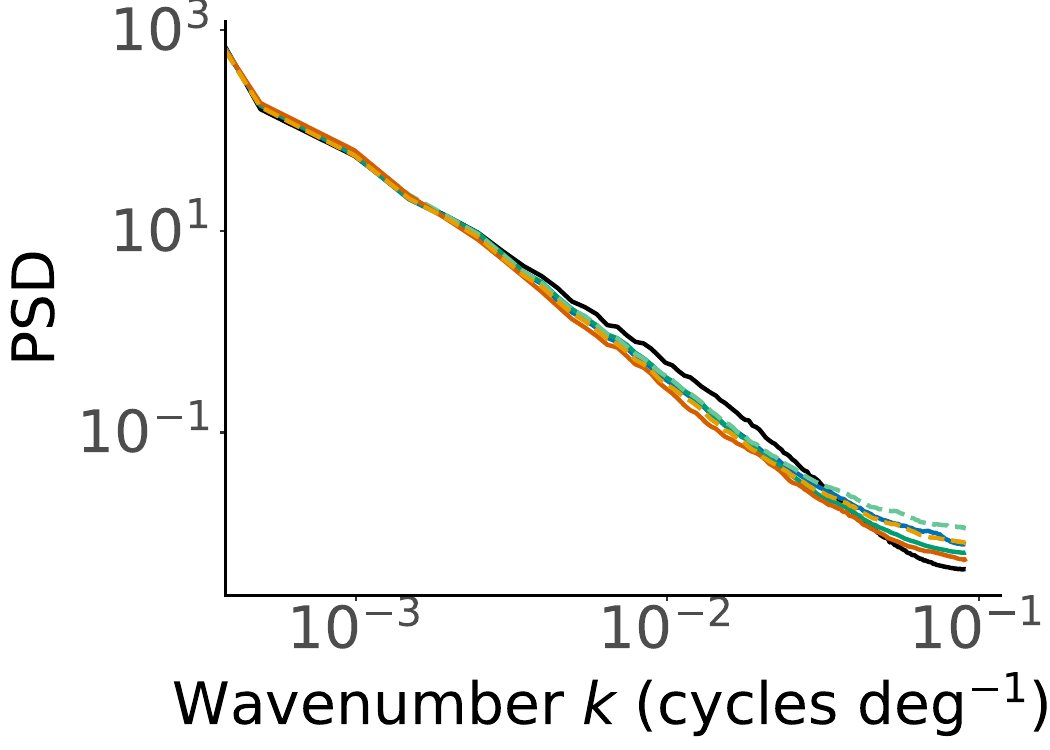}
    \caption{PSD of $\mathbf{E}_h$}
  \end{subfigure}

  \caption{PSD comparison of down-scaled and physics-derived variables from CERRA (ground truth) and model predictions in the Iberia, Morocco region (OOD). For all models, PSD curves deviate more from CERRA at high frequencies than in the in-distribution case, with an even larger gap for $\boldsymbol{\delta_h}$ and $\boldsymbol{\zeta_h}$.}
  \label{fig:psd_iberia_all}
\end{figure}

%%%%%%%%%%%%%%%%%%%%%%%%%%%%%%%%%%%%%%%%%%%%%%%%%%%%%%%%%%%%%%% SCANDINAVIA
\newpage
\section{Results Northern Scandinavia}
\label{app:results-Scandinavia}
%%%%%%%%%%% PLOTTED results - U wind - Iberia
\begin{figure}[htbp!]
  \centering
  % Row 1: two centered images
  \begin{subfigure}[htbp!]{0.23\textwidth}
    \centering
    \includegraphics[width=\linewidth]{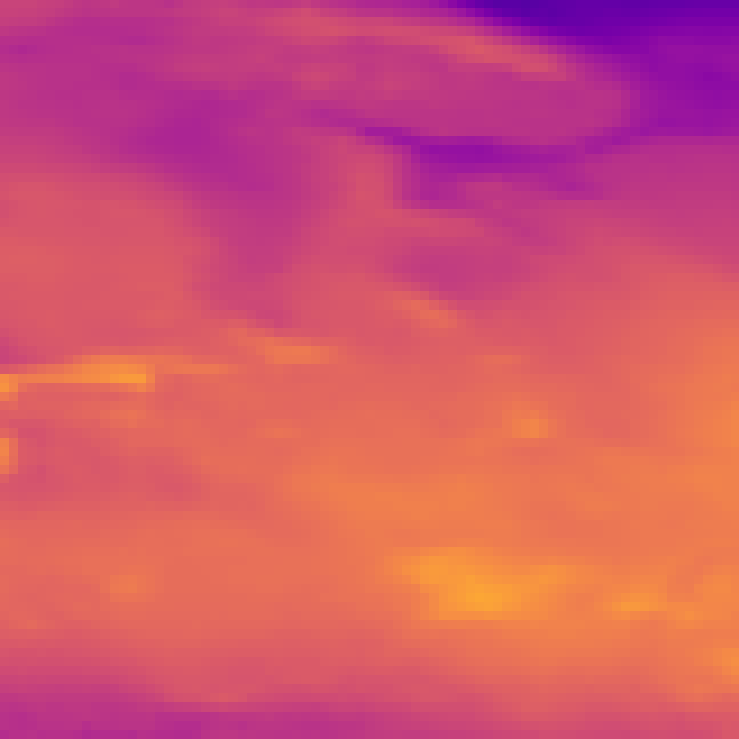}
    \subcaption{Input: ERA5}
    
  \end{subfigure}
  \hspace{0.04\textwidth} % small gap to center two images
  \begin{subfigure}[htbp!]{0.23\textwidth}
    \centering
    \includegraphics[width=\linewidth]{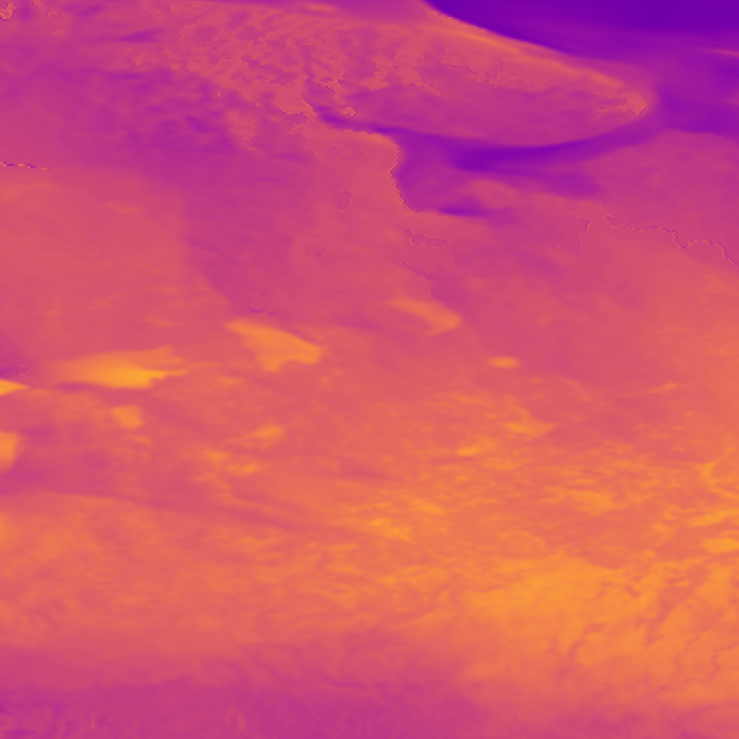}
    \subcaption{Target: CERRA}
    
  \end{subfigure}

  \vspace{0.3em}

  % Row 2: three equally spaced images
  \begin{subfigure}[t]{0.23\textwidth}
    \centering
    \includegraphics[width=\linewidth]{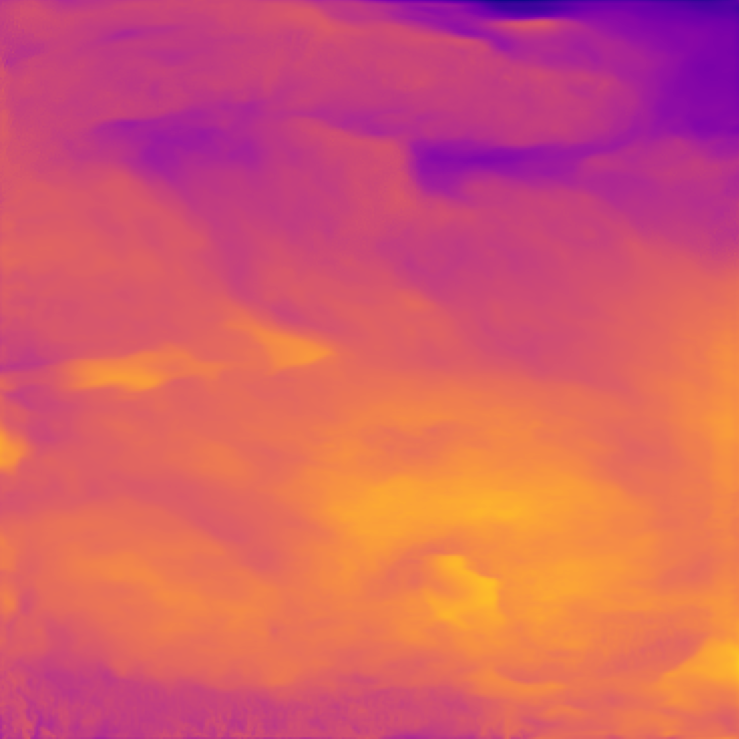}
    \subcaption{CorrDiff}
    
  \end{subfigure}
  \hfill
  \begin{subfigure}[t]{0.23\textwidth}
    \centering
    \includegraphics[width=\linewidth]{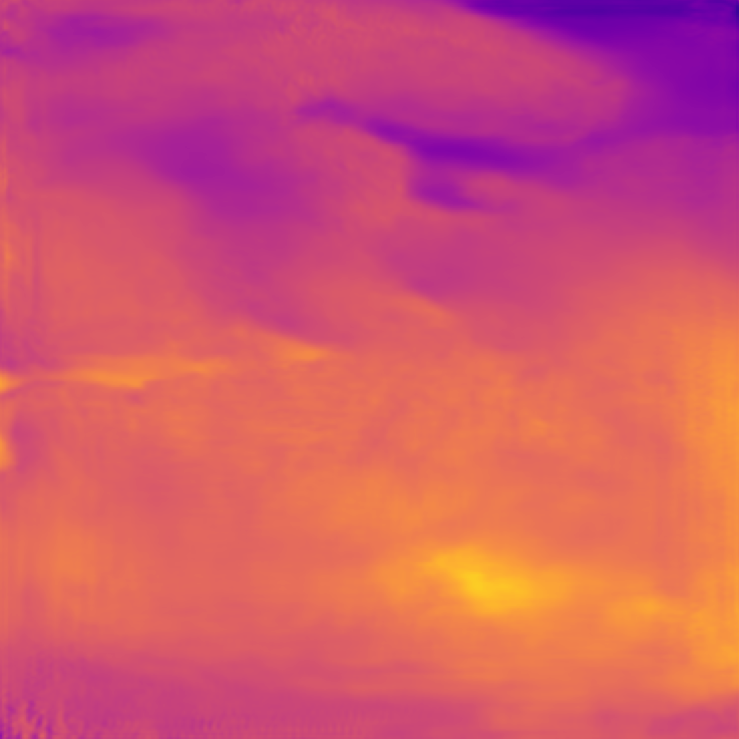}
    \subcaption{CRPS UNets-PSD}
    
  \end{subfigure}
  \hfill
  \begin{subfigure}[t]{0.23\textwidth}
    \centering
    \includegraphics[width=\linewidth]{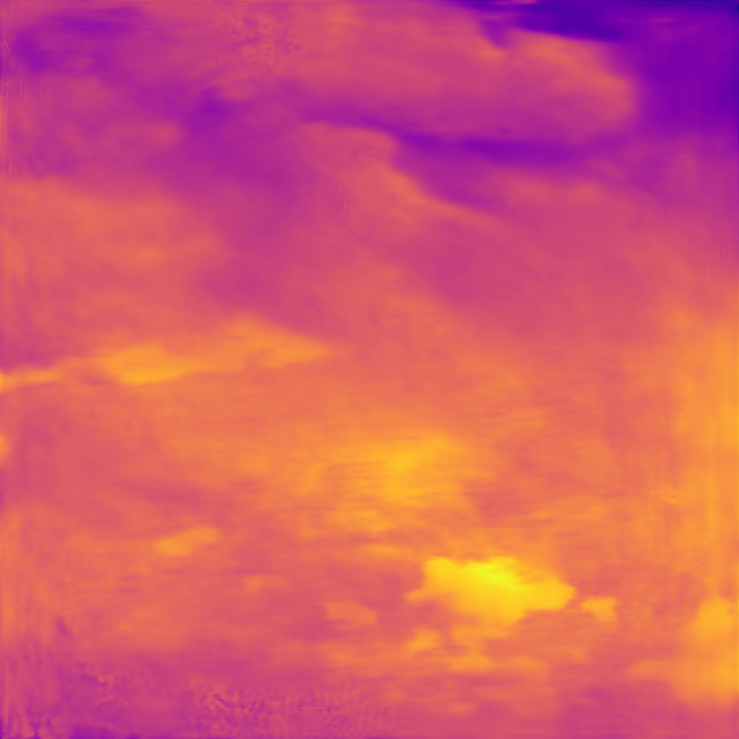}
    \subcaption{Reg-CorrDiff-PSD}
    
  \end{subfigure}

  \vspace{0.3em}

  % Row 3: full-width colormap
  \begin{subfigure}[t]{0.41\textwidth}
    \centering
    \includegraphics[width=\linewidth]{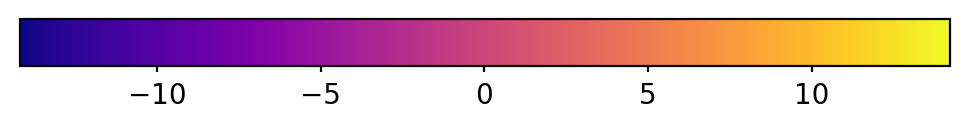}
    
  \end{subfigure}

   \vspace{0.3em}

  \caption{Input, target, and predictions of $\mathbf{u}$ in Northern Scandinavia (OOD) using \textit{CorrDiff}, \textit{Reg-CorrDiff-PSD}, and \textit{CRPS UNets-PSD}. Visually, all models miss many details of the target $\mathbf{u}$, making it difficult to see clear differences in quality. }
  \label{fig:qualitative-grid}
\end{figure}

%%%%%%%%%%% PLOTTED results - T2m wind - Scandinavia
\begin{figure}[htbp!]
  \centering
  % Row 1: two centered images
  \begin{subfigure}[htbp!]{0.23\textwidth}
    \centering
    \includegraphics[width=\linewidth]{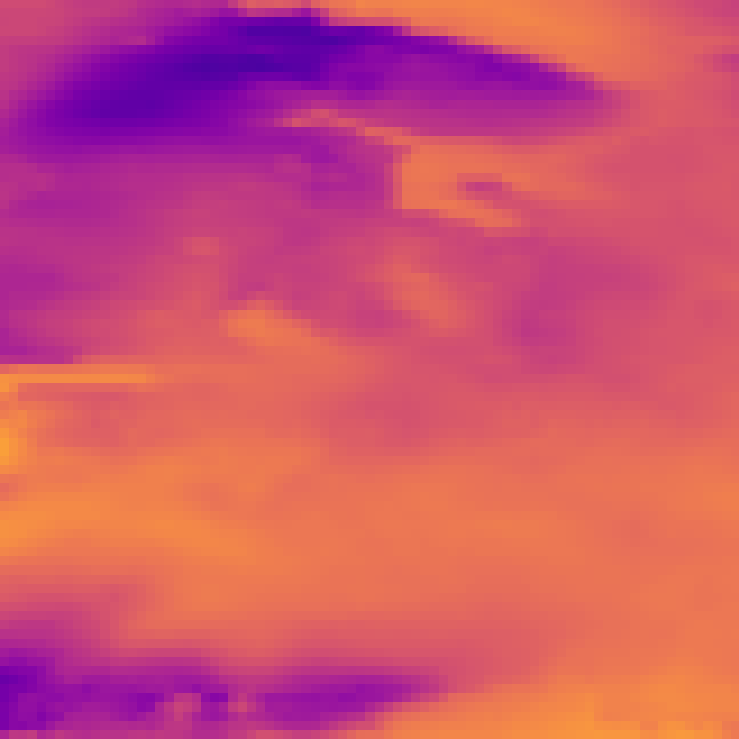}
    \subcaption{Input: ERA5}
    
  \end{subfigure}
  \hspace{0.04\textwidth} % small gap to center two images
  \begin{subfigure}[htbp!]{0.23\textwidth}
    \centering
    \includegraphics[width=\linewidth]{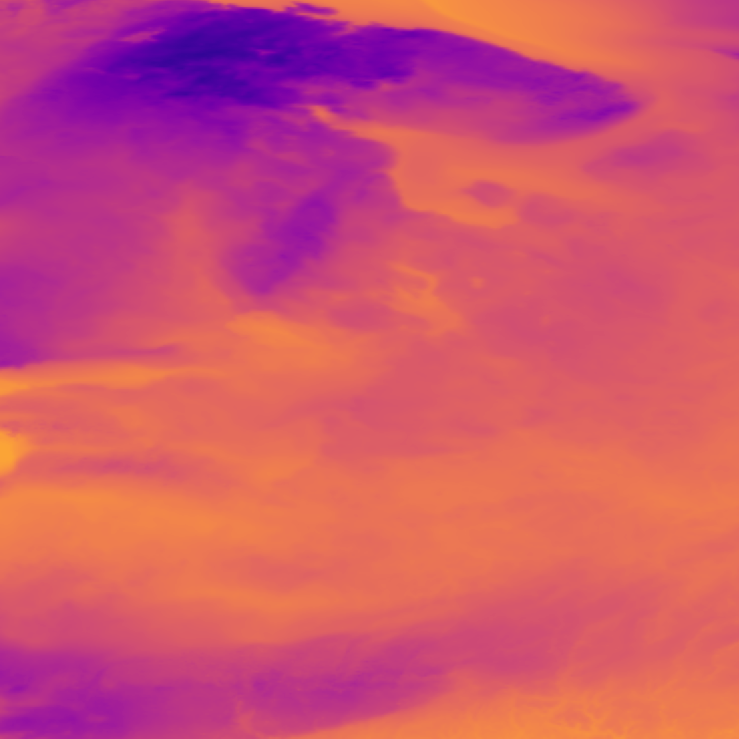}
    \subcaption{Target: CERRA}
    
  \end{subfigure}

  \vspace{0.3em}

  % Row 2: three equally spaced images
  \begin{subfigure}[t]{0.23\textwidth}
    \centering
    \includegraphics[width=\linewidth]{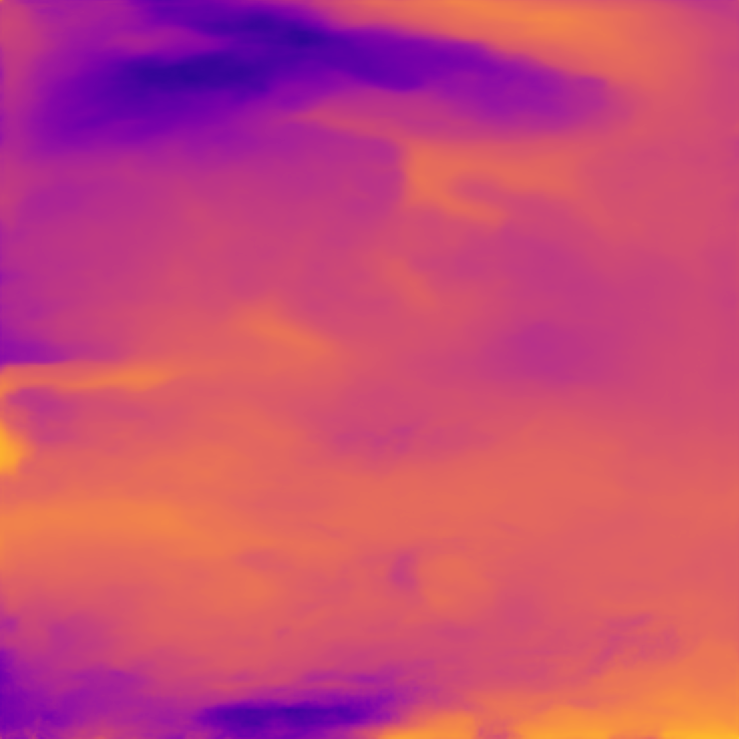}
    \subcaption{CorrDiff}
    
  \end{subfigure}
  \hfill
  \begin{subfigure}[t]{0.23\textwidth}
    \centering
    \includegraphics[width=\linewidth]{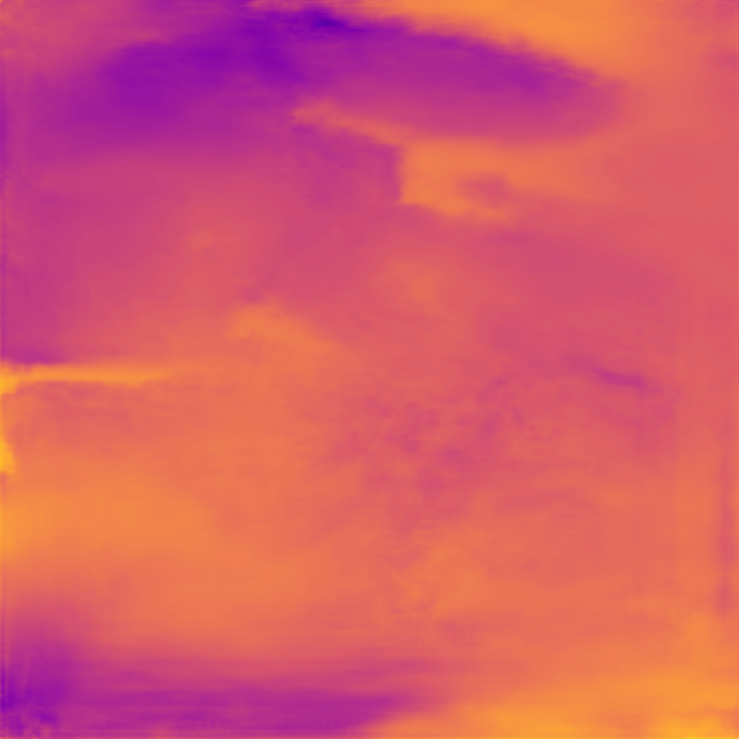}
    \subcaption{CRPS UNets-PSD}
    
  \end{subfigure}
  \hfill
  \begin{subfigure}[t]{0.23\textwidth}
    \centering
    \includegraphics[width=\linewidth]{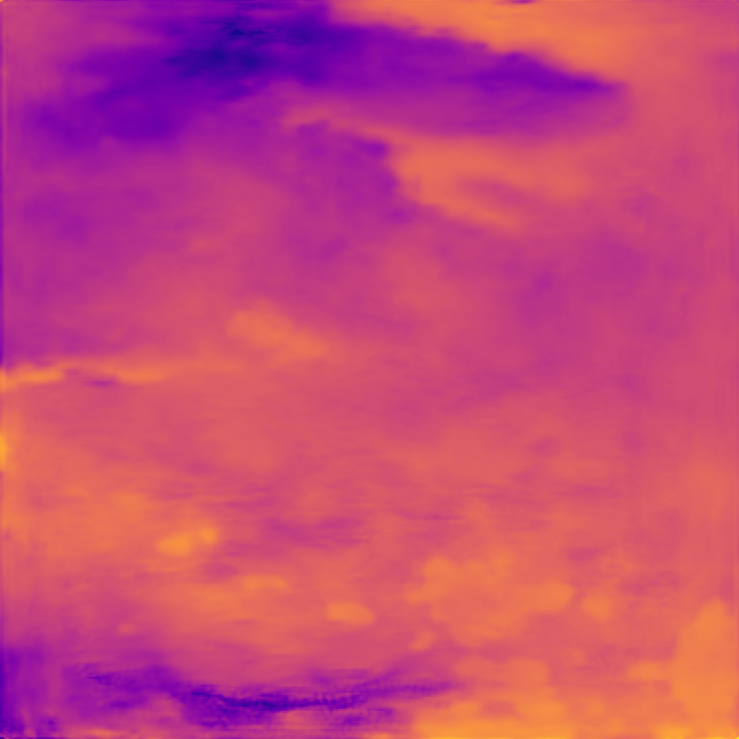}
    \subcaption{Reg-CorrDiff-PSD}
    
  \end{subfigure}

  \vspace{0.3em}

  % Row 3: full-width colormap
  \begin{subfigure}[t]{0.41\textwidth}
    \centering
    \includegraphics[width=\linewidth]{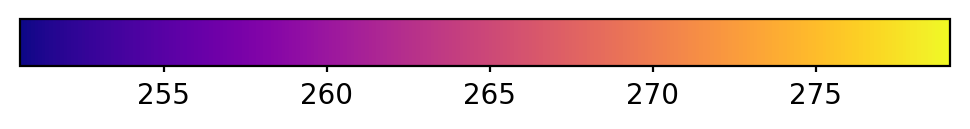}
    
  \end{subfigure}

   \vspace{0.3em}

  \caption{Input, target, and predictions of $\mathbf{t2m}$ in Northern Scandinavia (OOD) using \textit{CorrDiff}, \textit{Reg-CorrDiff-PSD}, and \textit{CRPS UNets-PSD}. Visually, all models miss many details of the target $\mathbf{t2m}$, making it difficult to see clear differences in quality.}
  \label{fig:qualitative-grid}
\end{figure}

%%%%%%%%%%% PSD Scandinavia, DOWNSCALED VARIABLES
\begin{figure}[htbp!]
  \centering
  \includegraphics[width=0.6\linewidth]{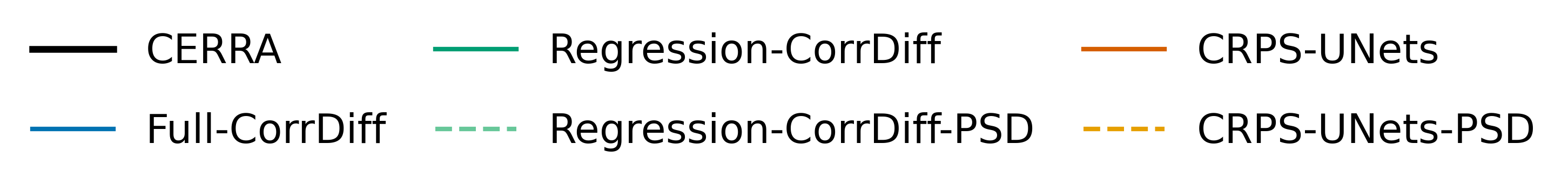}
\end{figure}
\vspace{-8pt}
%%%%%%%%%%% PSD IBERIA – ALL VARIABLES IN ONE FIGURE
\begin{figure}[htbp!]
  \centering
  %------------------ 1st ROW ------------------
  \begin{subfigure}[t]{0.31\textwidth}
    \centering
    \includegraphics[width=\linewidth]{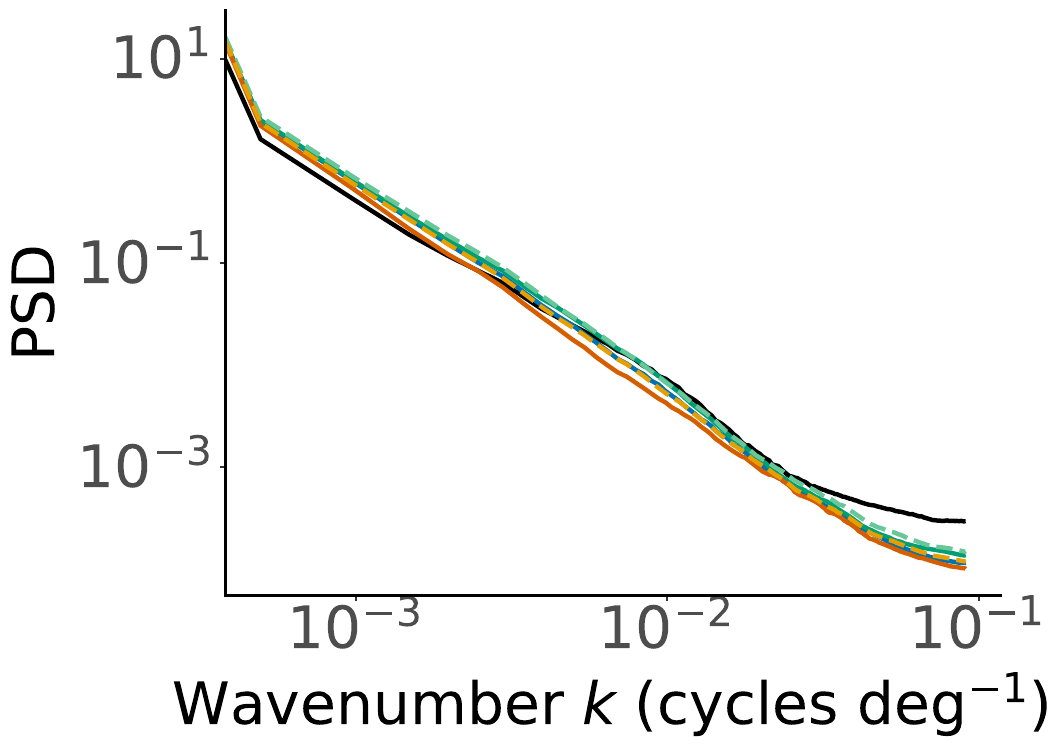}
    \caption{PSD of $\mathbf{u}$}
  \end{subfigure}\hfill
  \begin{subfigure}[t]{0.31\textwidth}
    \centering
    \includegraphics[width=\linewidth]{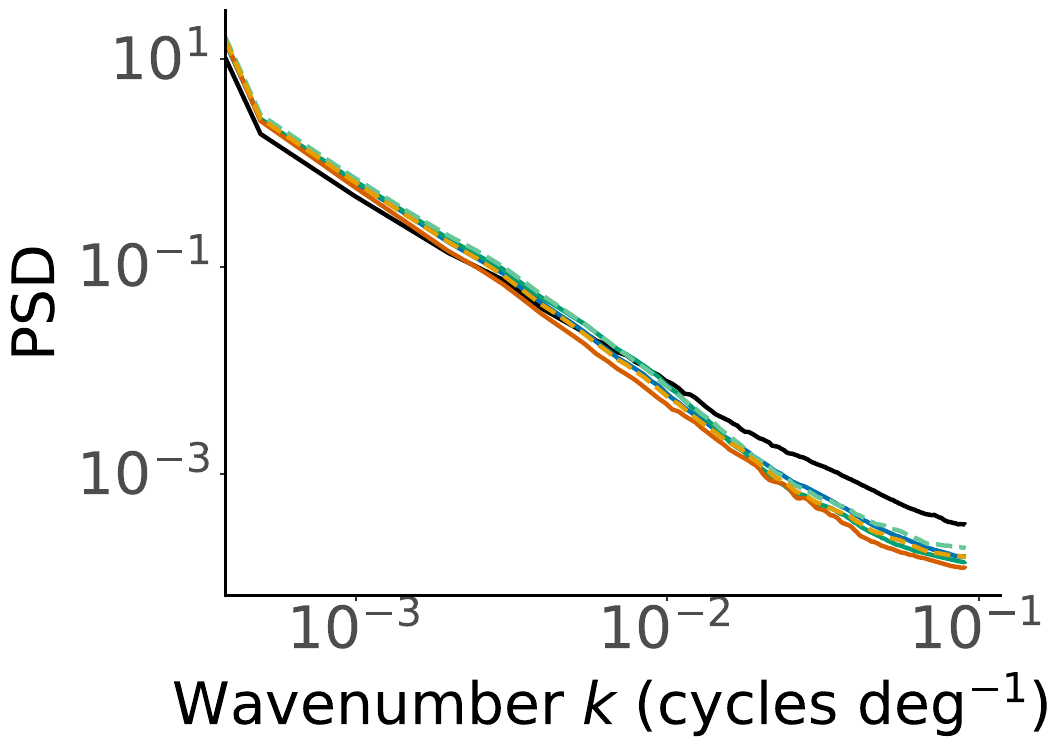}
    \caption{PSD of $\mathbf{v}$}
  \end{subfigure}\hfill
  \begin{subfigure}[t]{0.31\textwidth}
    \centering
    \includegraphics[width=\linewidth]{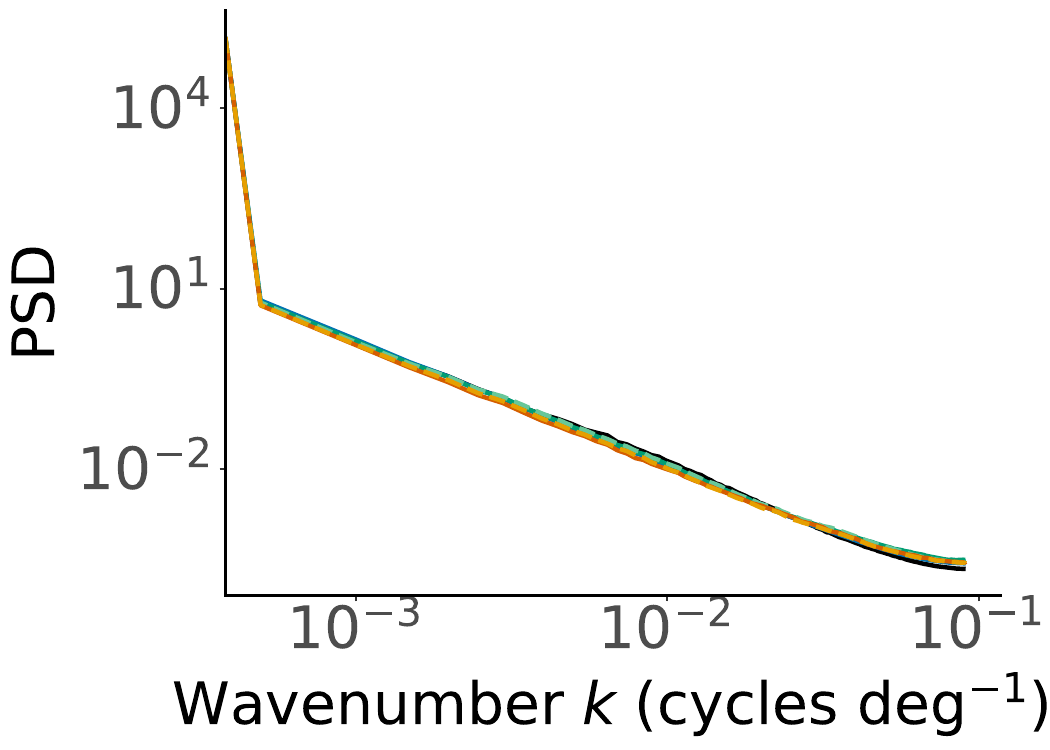}
    \caption{PSD of temperature $\mathbf{t2m}$}
  \end{subfigure}

  \vspace{0.5em} % vertical gap between rows
  
  %------------------ 2nd ROW ------------------
  \begin{subfigure}[t]{0.31\textwidth}
    \centering
    \includegraphics[width=\linewidth]{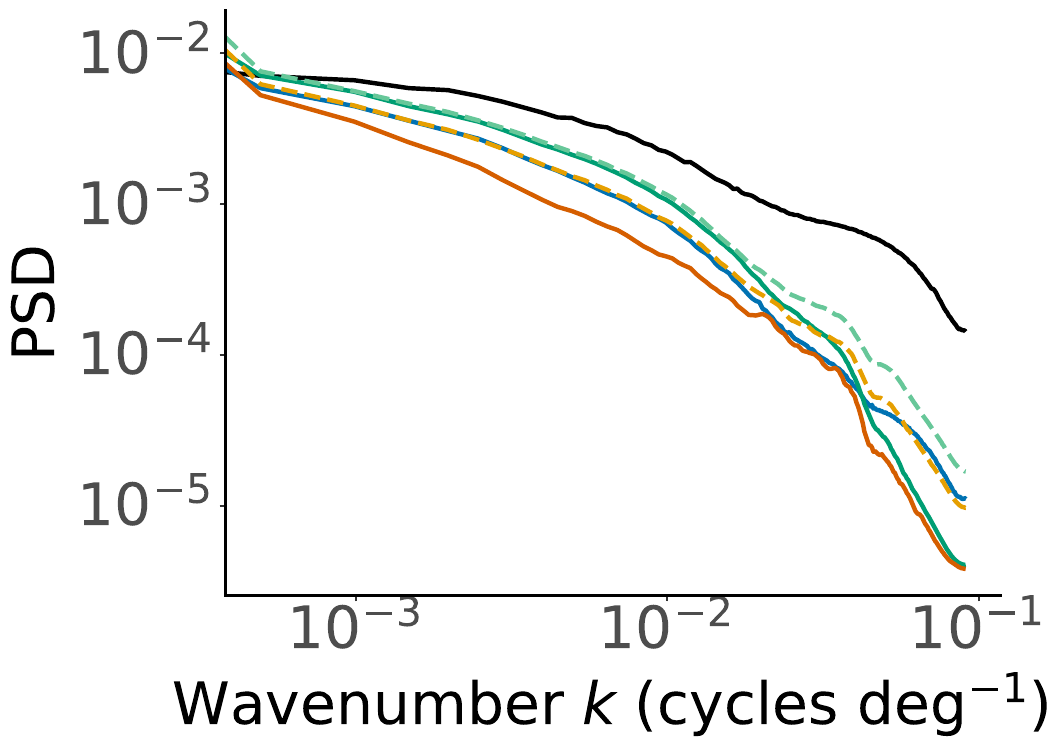}
    \caption{PSD of $\boldsymbol{\delta}_h$}
  \end{subfigure}\hfill
  \begin{subfigure}[t]{0.31\textwidth}
    \centering
    \includegraphics[width=\linewidth]{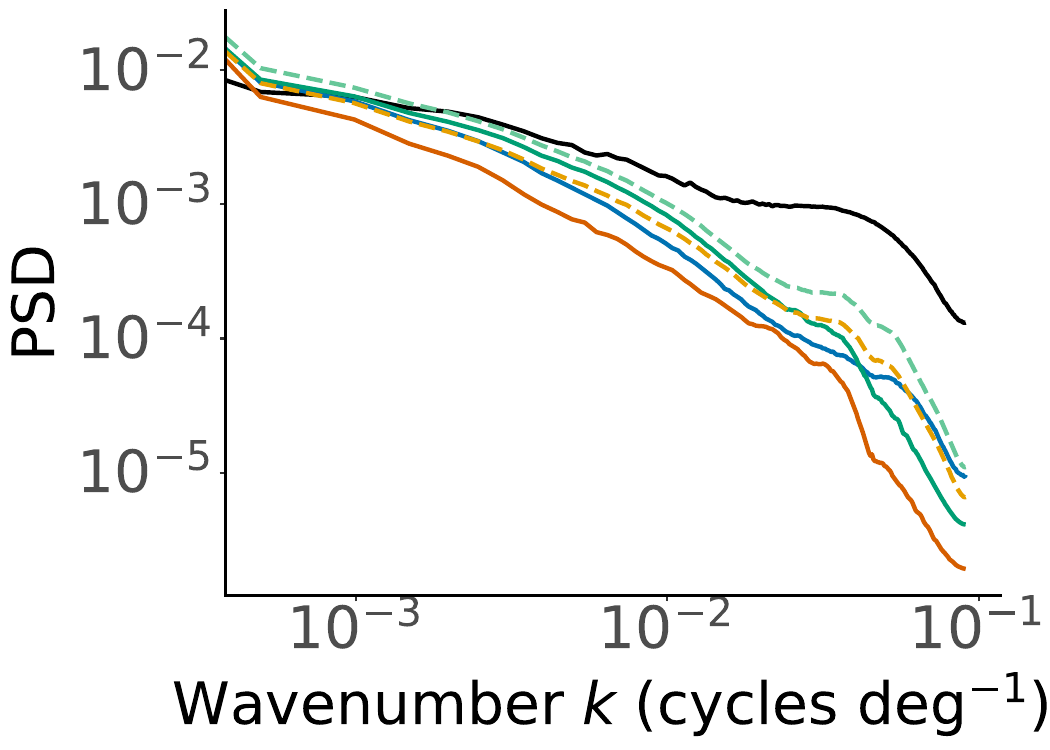}
    \caption{PSD of $\boldsymbol{\zeta}_h$}
  \end{subfigure}\hfill
  \begin{subfigure}[t]{0.31\textwidth}
    \centering
    \includegraphics[width=\linewidth]{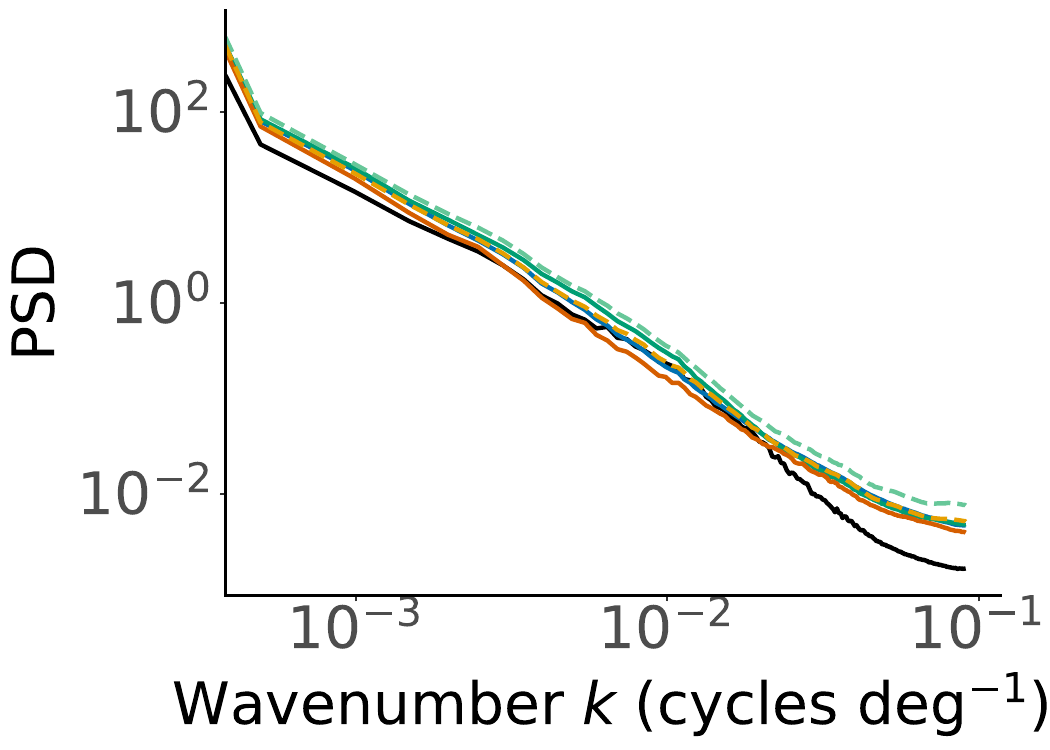}
    \caption{PSD of $\mathbf{E}_h$}
  \end{subfigure}

  \caption{PSD comparison of down-scaled and physics-derived variables from CERRA (ground truth) and model predictions in Northern Scandinavia (OOD). For all models, PSD curves deviate more from CERRA at high frequencies than in the in-distribution case, with an even larger gap for $\boldsymbol{\delta_h}$ and $\boldsymbol{\zeta_h}$.}
  \label{fig:psd_scandinavia_all}
\end{figure}

\end{document}